\documentclass{article}

\AtBeginDocument{%
  }

\usepackage[preprint,nonatbib]{neurips_2020}
\usepackage[utf8]{inputenc} % allow utf-8 input
\usepackage[T1]{fontenc}    % use 8-bit T1 fonts
\usepackage{hyperref}       % hyperlinks
\usepackage{url}            % simple URL typesetting
\usepackage{booktabs}       % professional-quality tables
\usepackage{amsfonts}       % blackboard math symbols
\usepackage{nicefrac}       % compact symbols for 1/2, etc.
\usepackage{microtype}      % microtypography
\usepackage{soul}
\usepackage{xcolor}
\usepackage{multirow}
\usepackage{hyperref}
\usepackage{mathtools}
\usepackage{todonotes}
\usepackage{caption}
\usepackage{subcaption}
\usepackage{amsmath}
\usepackage{fontawesome}
\usepackage{graphicx}
\usepackage{algorithm}
\usepackage{algpseudocode}
\usepackage[backend=biber]{biblatex}
\begin{document}

%\title{Curiosity in Policy Search for Spare Reward Environments}
\title{Curiosity creates Diversity in Policy Search}
%\title{Curiosity creates Diversity in Policy Search for Sparse Reward Environments}


\author{Paul-Antoine Le Tolguenec\\
  ISAE Supaero, Airbus\\
\texttt{paul-antoine.le-tolguenec@airbus.com}\\
\And
Emmanuel Rachelson\\
ISAE Supaero, Université de Toulouse\\
\texttt{emmanuel.rachelson@isae-supaero.fr}\\
\And
Yann Besse\\
Airbus\\
\texttt{yann.besse@airbus.com}\\
\And
Dennis G. Wilson\\
ISAE Supaero, Université de Toulouse\\
\texttt{dennis.wilson@isae-supaero.fr}}
%\renewcommand{\shortauthors}{Le Tolguenec et al.}

\maketitle

\begin{abstract}
    When searching for policies, reward-sparse environments often lack sufficient information about which behaviors to improve upon or avoid. In such environments, the policy search process is bound to blindly search for reward-yielding transitions and no early reward can bias this search in one direction or another. A way to overcome this is to use intrinsic motivation in order to explore new transitions until a reward is found. In this work, we use a recently proposed definition of intrinsic motivation, Curiosity, in an evolutionary policy search method. We propose Curiosity-ES \footnote{\href{https://github.com/SuReLI/Curiosity-ES}{https://github.com/SuReLI/Curiosity-ES}}, an evolutionary strategy adapted to use Curiosity as a fitness metric. We compare Curiosity-ES with other evolutionary algorithms intended for exploration, as well as with Curiosity-based reinforcement learning, and find that Curiosity-ES can generate higher diversity without the need for an explicit diversity criterion and leads to more policies which find reward.
\end{abstract}

\section{Introduction}

% Motivation
Finding the optimal policy in an environment is a trade-off between exploration and exploitation and early rewards can guide this trade-off towards rewarding parts of environment.
However, in reward-sparse environments, all rewards are null except in some specific goal states; finding an optimal policy therefore becomes a difficult and uninformed exploration problem.
Many real-world problems can be modelled as reward-sparse Markov Decision Processes (MDPs), such as robotic control \cite{hafner2019dream, andrychowicz2017hindsight} and autonomous vehicle stress testing \cite{koren2018adaptive}.

% Standard RL approaches for sparse
Many traditional reinforcement learning (RL) algorithms, such as \cite{mnih2015human}, handle this trade-off by adding some form of noise to the most promising action. Similarly, population-based policy search methods like evolutionary strategies (ES), such as \cite{salimans2017evolution}, generate a variety of exploratory control policies by introducing noise in the policy parameter space.
However, when the reward is sparse, noise-based approaches explore only a local part of the environment and fail to find rewarding states, for example in robotic control tasks \cite{andrychowicz2017hindsight}.

% Exploration on transitions
A separate approach to overcome the sparse reward problem is to use exploration bonuses to reward exploration. Early examples of exploration bonuses are count-based exploration methods \cite{brafman2002r} which count the number of visits to a state and reward visiting states which have low counts. More recent approaches use approximations of exploration with neural networks, such as Random Network Distillation (RND) \cite{burda2018exploration} which uses the prediction error between a randomly initialized network and a distilled network as the exploration bonus. Variational Intrinsic Control \cite{gregor2016variational} and the Diversity is All You Need algorithm (DIAYN) \cite{eysenbach2018diversity} use measures from information theory to quantify the distribution of states visited by an RL policy, which is used in an exploration reward. In this work, we specifically focus on intrinsic motivation methods \cite{aubret2019survey} which reward policies for covering unexplored or under-explored transitions. We base this work on Curiosity \cite{pathak2017curiosity}, a type of intrinsic motivation which rewards policies for discovering transitions that are different from those seen previously.

% QD on states
Population-based methods have been used on sparse reward environments, specifically through the use of objective functions which encourage exploration. Quality Diversity (QD) methods, such as Novelty Search (NS) \cite{lehman2011abandoning} and MAP-Elites \cite{cully_robots_2015}, are Genetic Algorithms that encourage policies to cover new areas of a predefined behavior space, such as the terminal state of a policy. Novelty has been demonstrated to be a useful intrinsic motivation for ES \cite{conti2018improving} but requires the definition of a behavior space, which can be a limiting factor in applying QD methods.

% Our contribution
In this paper, we show that Curiosity can be used as an intrinsic fitness metric to explore sparse reward environments. We use Curiosity as a self-supervised prediction metric on transitions covered throughout the full episode which accumulates to an overall Curiosity fitness score. We study Curiosity as the intrinsic fitness of an evolutionary strategy by proposing Curiosity-ES. We demonstrate that Curiosity-ES outperforms state of the art QD methods such as MAP-Elites on maze navigation and robotic control tasks without requiring an explicit behavior definition. We show empirically that Curiosity leads to the exploration of transitions which are different from previously seen transitions, allowing for multiple reward-finding trajectories to be discovered.

%it is possible to explore sparse or deceptive environments using a Curiosity mechanism. 
%Since this method does not require the creation of a behavior, we show that our approach can compete with NS on maze benchmarks and achieve better exploration in other environments.

% Layout of the paper
We first contextualize Curiosity-ES with other policy search algorithms, notably Novelty Search and other Evolutionary Strategies, in \autoref{sec:background}. We expand on Curiosity and its use in Reinforcement Learning in \autoref{sec:curiosity}. We then present Curiosity-ES in \autoref{sec:curiosity-es}, detailing how Curiosity is calculated over a trajectory and aggregated as a fitness metric. The maze navigation and robotic control environments and experimental parameters used in this article are detailed \autoref{sec:experiments}. We compare Curiosity-ES to a set of QD and RL algorithms on these two environments using both extrinsic and intrinsic reward in \autoref{sec:reward}, finding that Curiosity leads to more exploration and more efficient policies on both types of environments. We provide a detailed study of the diversity of rewarding policies on of the maze navigation tasks in \autoref{sec:diversity_rewarding_policy}. Finally, we discuss the implications of Curiosity in population-based methods and define future directions in \autoref{sec:discussion}.

% One needs to explore the environment in order to find new reward-yielding transitions and, at the same time, one needs to explore the data collected to bias the algorithm toward the optimal parameters.
% In reward-sparse environments, finding the optimal policy that is to say the policy that leads to specific state(s) becomes a hard exploration problem. 
% At any time step during its interaction with the environment, the RL agent needs to choose between following the action it currently believes sets it on the most rewarding course through the state space, to better explore later around this course, or conversely to take an immediate exploratory action in hope to discover even better trajectories.
%Such rare rewards provided by the environment are often coined \emph{extrinsic rewards}.
% MDP-Based algorithms such as Deep-Q Learning or Policy gradient methods usually explore the environment by noising the action space. 
% Evolutionary strategies explore the environment by generating individuals with different parameters that is to say by noising the parameter space.
% The trade-off between exploring new transitions and exploiting the current best policy should guarantee that, asymptotically, all transitions in the environment are explored and, at the same time, that the necessary data to identify an optimal policy is found early. 
%When the rewards are sparse, this is likely to lead to inefficient and local exploration around the starting state.

\section{Background}
\label{sec:background}

Policies which explore rather than exploit have been encouraged by various mechanisms in the policy search and RL literature. Population-based methods are by their nature exploratory through their use of random modifications to existing policies, but divergent search methods such as Novelty Search and Quality Diversity go further to search explicitly for a diversity of behaviors over entire episodes. RL methods such as intrinsic motivation, on the other hand, often add loss terms which encourage diversity on individual transitions by taking random actions or for having discovered a new state. Curiosity-ES is inspired by both approaches, using transition-level exploration bonuses in a population of agents.

\subsection{Novelty Search and Quality Diversity}

% definition of novelty: behavior space, distance in BS
Population-based algorithms such as Genetic Algorithms \cite{holland1992genetic} simulate natural selection over a population of individuals in order to maximize an objective (fitness) function, usually resulting in a set of optimal solutions.
However, natural evolution diverges, creating and maintaining a wide variety of solutions for different problems.
Novelty Search \cite{lehman2011abandoning} and other Quality Diversity algorithms \cite{chatzilygeroudis2021quality} use a behavior characterisation for each generated individual in order to maintain behavioral diversity.
Behaviors are characterized by a function $b$ that takes in the individual parameter vector (genome) $\theta_{i}$ and output its corresponding behavior $b_{\theta_{i}}$, often based on the trajectory taken in an environment. The distance between the behavior of individuals is used to bias search towards exploration into new parts of the behavior space.

% diversity algorithms: QD, OEE
%NS is an approach that belongs to the family of open-ended evolution.
Quality Diversity algorithms have been demonstrated on a number of domains, such as robotic control \cite{cully_robots_2015}, urban design \cite{galanos2021arch}, and chair and lamp design \cite{xu2012fit}. In these methods, behavior can be used to calculate an intrinsic fitness based on distance to existing individuals, as in NS-ES \cite{conti2018improving}, or to store individuals with a diversity of behaviors, as in MAP-Elites \cite{cully_robots_2015}. The use of behavioral diversity as a search mechanism has made them especially suited for reward-sparse environments such as the Atari games Pitfall and Montezuma's revenge \cite{ecoffet2021first}. Novelty Search in particular has often been studied using robotic locomotion or maze navigation tasks \cite{lehman2011abandoning, lehman2011evolving, conti2018improving}, where the final state of an agent is used as the behavior descriptor. Searching for diversity in the final state of the agent therefore brings it to different states, some of which may be finally rewarding. 

QD algorithms aim to optimize both exploration and an extrinsic fitness, often the sum of reward of a policy. As shown in the next section, NS-ES \cite{conti2018improving} does so using a weighted sum from an exploration fitness term and an objective fitness term, as we do in Curiosity-ES. MAP-Elites \cite{cully_robots_2015} and Novelty Search with Local Competition (NSLC) \cite{lehman2011evolving} balance quality and diversity by simulating competition, based on extrinsic fitness, between policies that have similar behavior. Covariance Matrix Adaptation MAP-Elite (CMAME) \cite{fontaine2020covariance}, which is based on MAP-Elites, balances exploration and exploitation in the same was as MAP-Elites, but uses an ES for policy optimization.

% problem of behavior descriptor: TAXONS, AURORA
% TAXONS and AURORA uses neural network to create the space
The design of a useful behavior descriptor can often be a limiting factor for the use of QD algorithms. Recent methods such as AURORA \cite{cully_autonomous_2019,grillotti_unsupervised_2022} and TAXONS \cite{paolo2020unsupervised} propose the use of a neural network to learn a behavior descriptor throughout evolution. Similarly, we propose the use of a neural network to calculate Curiosity throughout evolution, however we do not rely on a behavior descriptor to calculate Novelty or behavioral diversity but rather directly compute an intrinsic fitness using Curiosity.

\subsection{Evolution Strategies for Policy Search}

% Evolutionary Strategies
  % competitive results: massive parallelization leading to small wall clock time convergence
  
Evolution strategies \cite{rechenberg1978evolutionsstrategien} have recently been shown to be competitive methods for policy search, even on difficult visual problems such as video games \cite{salimans2017evolution}. ES have the advantage of parallelization, leading to small wall clock time for evolution; for example, in \cite{tang2022evojax}, a hard version of the classic CartPole problem is solved in under 2 minutes on a single GPU. In this article, we base our ES on the Canonical ES presented in \cite{chrabaszcz2018back}, which achieves human-competitive results on the Atari benchmark in under an hour with a simple ES.

 % ES in detail
 
In an evolutionary strategy, a population of individuals $\theta$ is sampled from a given distribution; in policy search, these individuals correspond to function parameter vectors. Each individual is evaluated according to a given objective function $F : \mathbb{R}^{n} \to \mathbb{R}$. The individuals are then selected according to a specific rule and the sampling distribution is updated using the best individuals. Most ES differ in the sampling distribution update; while many use fixed Gaussian distributions, as we do in this work, adaptive methods like the Covariance Matrix Adaptation ES \cite{hansen2001completely} modify the distribution over search.

For each generation $k$, a population of $\lambda$ individuals is sampled : $\theta_{k}^{i} \sim \mathcal{N}(\theta_{k},\sigma)$ where $\sigma$ is a fixed parameter in Canonical ES and $\theta_{k}$ is the center of the generation's population. Each individual is evaluated using $f$: $f_{\theta_{k}^{i}} = f(\theta_{k}^{i})$. In many ES, including Canonical and CMA-ES, the individuals are then ranked according to their fitness in order to approximate the gradient ascent for the distribution center. In Canonical ES, individuals are sorted into $\{\mathcal{R}^{j}|j \in [0,\mu]\}=sort((\theta_{k}^{i},f_{\theta_{k}^{i}}), i \in [0,\mu])$, where $\mu \leq \lambda$. An estimation of the gradient is computed by weighting the $\mu$ best $\theta_{k}^{j}$ with a weight $w_{j}$ corresponding to their rank, and the center $\theta_k$ is updated by: 

\begin{equation}
\nabla_{\theta_{k}}\mathop{\mathbb{E}_{\theta_{k}^{i}\sim \mathcal{N}(\theta_{k},\sigma)}}[f(\theta_{k})]=\frac{1}{\sigma\mu}\sum_{j=0}^{\mu} (\mathcal{R}^{j}-\theta_{k})w_{j}
\end{equation}

% Objective function in policy search
In policy search, individuals represent functions which take actions in an environment and the objective function is the sum of reward gained over an episode. Specifically, we represent a policy function $\pi_{\theta}$ as a neural network with $\theta$ as parameters. This function takes in an environment state $s_{t}$ and returns an action $a_{t} = \pi_{\theta}(s_t)$. Based on the action, the environment then returns a reward $r_{t}$, which in the sparse case is often 0, according to some function $R(s_t, a_t)$. The environment then advances to the next state $s_{t+1}$ until a termination criterion is met, reaching a maximum timestep $T$. The objective function can therefore be defined as
\begin{equation}
    f(\theta)=\sum_{t=0}^{T} r_{t}=\sum_{t=0}^{T} R(s_{t}, \pi_{\theta}(s_{t})).
\end{equation}

%\subsection{Novelty Search Evolutionary Strategy}

Recently, evolutionary strategies have been combined with Novelty Search in \cite{conti2018improving}, specifically in the NS-ES algorithm. In NS-ES, the Novelty score $N$ is defined for a given $\theta_{i}$ as the sum of the distances between the individual behavior $b(\theta_{i})$ and the k-nearest neighbours ($kNN$) of that behavior in an archive $A$ of behaviors. The Novelty score $N$ is then used as a part of the fitness function $f$:
\begin{align}
K &= kNN(b(\theta_{i}),A) \\
N(\theta_{i},A) &= \frac{1}{|K|}\sum_{j\in K} \parallel b(\theta_{i})-b(\theta_{j}) \parallel_2 \\
    f(\theta) &= \varphi\sum_{t=0}^{T} r_{t} + (1-\varphi) N(\theta_{i},A)
\end{align}

where $\varphi \in [0,1]$ determines the ratio between reward-based fitness and Novelty. \cite{conti2018improving} propose other modifications to the ES such as a meta-population for sampling, but we focus on the use of Novelty as intrinsic motivation for comparison with Curiosity.

%For every individual evaluated during the evolution, the behavior $b_{\theta_{i}}$ is saved in an archive $A$. The behavior is used to move away from the closest behaviors of the archive and thus create diversity.
%(cite : Novelty search and the problem with objectives)

% NS creates an exploration bonus that pushes the evolution to sample new behaviors. As in RL problems the behavior is usually defined as the last state of the agent encountered in the environment, NS pushes the evolutionary schema to visit new states.
%The definition of a distance measure can be limiting in some settings. For example when the state of the agent is an image, the $\ell_2$ norm is no longer relevant. 

\subsection{Curiosity as intrinsic motivation}
\label{sec:curiosity}

% Curiosity
  % describe uses in the literature
  
In RL, intrinsic motivation has been proposed to encourage exploration of new transitions \cite{aubret2019survey}. One such intrinsic motivation is Curiosity through self-supervised prediction \cite{pathak2017curiosity}. This method uses an Intrinsic Curiosity Module (ICM) to produce intrinsic rewards based on prediction error of transitions. The goal of an ICM is to predict the next state $s_{t+1}$ based on the current state $s_t$ and the action taken in the environment $a_t$. The intrinsic reward is defined as the error between the prediction of next state from the forward model and the real new state provided by the environment.

The ICM is composed of three neural networks: an encoder $\phi_{w_e}$, a forward model $F_{w_f}$, and an inverse model $I_{w_i}$ with parameters $w_e$, $w_f$, and $w_i$ respectively. The Encoder $\phi_{w_e}$ maps the state space $S$ into a feature space $\mathcal{Z}$ ($\phi_{w_e}:S\to\mathcal{Z}$). This allows Curiosity to be calculated on large states such as images without the need for a hand-designed feature space. Prediction using the encoding of states $\phi_{w_e}(s_{t})$ is then used to train the ICM with loss terms depending on the forward and inverse models.

The forward model $F$ aims to predict the features of the next state $\phi_{w_e}(s_{t+1})$ based on the features of the current state $\phi_{w_e}(s_{t})$ and the action taken $a_{t}$ ($F_{w_{f}}:\mathcal{Z}\times A\to\mathcal{Z} $). The forward model loss is therefore:
\begin{equation}
    L_F(w_{f},w_{e})=\parallel F(\phi_{w_e}(s_{t}),a_{t})-\phi_{w_e}(s_{t+1})\parallel_{2}. 
\end{equation}

The inverse model is used to avoid representation collapse of the encoder model by predicting the action taken based on the feature space representation of the state and the next state ($I_{w_{i}}:\mathcal{Z}^{2}\to A$). The inverse loss corresponds to the error between the action $\hat{a}_{t}=I_{w_{i}}(\phi_{w_e}(s_t),\phi_{w_e}(s_{t+1}))$ predicted by $I_{w_{i}}$ and the actual action $a_t$ experienced in the environment to transition from $s_{t}$ and $s_{t+1}$. 
\begin{equation}
L_{I}(w_{i},w_{e})=\parallel I_{w_{i}}(\phi_{w_e}(s_t),\phi_{w_e}(s_{t+1}))-a_t \parallel_2.
\end{equation}

We note that we use the $\ell_2$ norm for $L_I$ as we use continuous actions, where \cite{pathak2017curiosity} used discrete actions and therefore a softmax loss.
The optimization of the ICM is a combination of the two loss terms, simultaneously used to train all three networks
\begin{equation}
    L_{ICM}(w_{i},w_{f},w_{e})=(1-\beta)L_{I}+\beta L_{F},
\end{equation}
where $\beta \in [0,1]$ is a parameter that weights the inverse model loss against the forward model. This parameter is important as training the feature encoder using only the forward model loss can lead to representation collapse of an encoder which maps any state $s_{t}$ to $\{0\}_{\mathcal{Z}}$.

As the forward model becomes more and more accurate in predicting $\phi_{w_e}(s_{t+1})$ given $\phi_{w_e}(s_{t})$ and $a_{t}$ for explored area of the environment, the intrinsic reward (e.g the exploration bonus) can be defined as: 
\begin{equation}
    r_{t}^{i}=\frac{\eta}{2}\parallel F_{w_f}(\phi_{w_e}(s_{t}),a_{t})-\phi_{w_e}(s_{t+1})\parallel_{2}
\end{equation}
where $\eta$ is a parameter that weight $r_{t}^{i}$. In other words, the reward will be high when the forward model's accuracy is low, suggesting that the forward model has not yet encountered that transition. The reward is not computed using the inverse loss model because the exact action taken can sometimes be impossible to predict, for example in a constrained MDP like a maze where multiple actions can lead to the same state.

In \cite{pathak2017curiosity}, the ICM is frequently updated during RL training to minimize the error between the prediction of the forward model and the actual new state encountered. Therefore, the reward is high in the non-visited areas of the environment, i.e. the regions in which the forward model has not yet been trained to convergence. In Curiosity-ES, this allows for the use of samples generated by the entire population to direct search towards under-explored transitions.

% Paragraphe on the added value of Curiosity-ES with respect to Curiosity-RL 

Our motivation for adapting Curiosity as an intrinsic motivation from RL to ES-based policy search is two-fold: we believe that the parameter-noising exploration of an ES is better adapted to the adversarial nature of the ICM training and that the use of a population of policies will cover new areas more exhaustively. We expand on both ideas briefly.

The ICM as defined above is an adversarial schema. The ICM attempts to learn the dynamics of the environment while the policy optimization tries to maximize ICM error by exploring transitions which the ICM is not capable of reconstructing. This adversarial optimization can be written as:
\begin{equation}
    \min_{w_f,w_e}\max_{\pi} \mathbb{E}_{(s_t,a_t,s_{t+1})\sim\rho_{\pi}}\parallel F_{w_f}(\phi_{w_e}(s_{t}),a_{t})-\phi_{w_e}(s_{t+1})\parallel_{2}.
\end{equation}

% In such a setting, exploration concentrates around the equilibrium. If the ICM fails to learn the dynamic of the environment, there will be continuously high Curiosity without requiring exploration. In practice, this may happen in environments with high transition entropy or in partially observable environments. For deterministic MDPs, however, we expect the ICM training to converge.

When the ICM learns faster than the policy search explores, then no exploration bonuses will be produced as long as the policy continues to sample similar transitions to those already observed. In such a case, exploration will depend solely on the exploration mechanism of the policy optimization method, which is often the use of random actions in RL algorithms. We believe that exploration based on network parameter modification, as done in ES, will lead more effectively to sampling new transitions than the pseudo-random policies of RL.

Our second point posits that a population of similar agents, as produced in a ES generation, will explore more comprehensively different areas of the transition space of an MDP. Curiosity is by definition a consumable resource \cite{ecoffet2019go}; in other words, once the ICM has learned the dynamics of a set of transitions, exploration will be encouraged in new parts of the environment's transition space. A single agent learning in this new area may only observe a subset of the possible transitions before the ICM has again trained on the new area. By using a population of policies to explore a new area between training steps of the ICM, we believe that new areas will be more fully explored.

While Curiosity can bring advantageous exploration in a reinforcement learning algorithm, the inherent exploration of ES can further benefit the adversarial training of the ICM. In \autoref{sec:reward}, we compare Curiosity-ES to the Twin Delayed Deep Deterministic (TD3) policy gradient algorithm \cite{fujimoto2018addressing} with intrinsic reward from the ICM to illustrate this point.

%The comparison is made in the feature space instead of the state space to allow the use of the $\ell_2$ norm which would be impracticable for state space like images. 
%Instead of hand-designing a feature space for each environment, ICM use general rules that the feature space must respect through the training.
%Let denote $F_{w_{f}}:\mathcal{Z}\times A\to\mathcal{Z} $, $I_{w_{i}}:\mathcal{Z}^{2}\to A$ and $\phi_{w_e}:S\to\mathcal{Z}$ respectively the forward dynamic model, the inverse dynamic model and the encoder that maps the state space to the feature space $\mathcal{Z}$. 
%$F$ and $I$ are defined in the feature space. 
%$\phi$ is trained in order to minimize two loss function. 
%The first one is the inverse loss that define the prediction error of the inverse model $I_{w_{i}}$.
%The inverse loss correspond to the error between the action $\hat{a}_{t}=I_{w_{i}}(\phi(s_t),\phi(s_{t+1}))$ predicted by $I_{w_{i}}$ and the actual action experienced in the environment to transition from $s_{t}$ and $s_{t+1}$:
%$$L_{I}(w_{i},w_{e})=\parallel I_{w_{i}}(\phi(s_t),\phi(s_{t+1}))-a_t \parallel_2$$
%The second loss, correspond to the error between the prediction of the forward model $\hat{\phi}(s_{t+1})=F(\phi(s_{t}),a_{t})$ and the feature space representation of the next state $\phi(s_{t+1})$ :
%For some state space making a prediction in the state space would be too noisy so the prediction and the comparison are made in the feature space. This slight difference allows to use ICM in environments with image state representation. 

\section{Curiosity-ES}
\label{sec:curiosity-es}

% short overview of how Curiosity-ES works, no math

%When trying to solve sparse reward environments, a good idea is to generate a maximum of data. Indeed, without even mentioning the method we use to explore hard-to-reach area of the environment, visiting different states is in itself a good strategy since it maximizes the probability of seeing rewarding states.It is one of the reasons that makes evolutionary strategies a tool of choice for the search of optimal policy in a sparse environment.Indeed, one of the reasons why evolutionary strategies were used in the first place is because these algorithms parallelize better.
%However, In some environments, the probability of reaching rewarding states is extremely low and a standard ES may not be able to reach them in a reasonable time. Hence, we must use an Intrinsic motivation to drive the exploration to reach unreachable states. 
%For this work we have used evolutionary strategies guided by an intrinsic reward produced by Curiosity.

We propose the use of Curiosity as intrinsic motivation for policy search in an evolutionary strategy through Curiosity-ES. Following a standard ES, policies represented as the continuous parameters of a neural network are generated by randomly sampling from a distribution. These networks are then evaluated on sequential decision tasks and the sum of reward over an episode is used as extrinsic fitness to estimate a gradient update for the population distribution. In Curiosity-ES, we also use an ICM to compute the Curiosity over each individual's entire trajectory. The Curiosity is then aggregated as an intrinsic fitness term which is added to the extrinsic fitness for the gradient estimation. A subset of the transitions are added to a replay buffer for training the ICM, which is done once per generation. The pseudocode of Curiosity-ES is presented in \autoref{alg:curiosity} and detailed below.

\begin{algorithm}[!ht]
\caption{Curiosity-ES}\label{alg:curiosity}
\begin{algorithmic}
\Require $\mu, \lambda, \sigma, \alpha, \alpha_{ICM}, \beta, \gamma, m, p, N, \varphi, w_j $
\State Initialize : $\theta_0, w_f, w_i, w_e, \mathcal{D}$
% \State $y \gets 1$
% \State $X \gets x$
% \State $N \gets n$
\For{$k=1,N$} \Comment{$N$ generations}
    \For{$i=1,\lambda$} \Comment{$\lambda$ individuals}
        \State $\theta_{k}^{i } \sim \mathcal{N}(\theta_{k},I\sigma)$ \Comment{sample individual}
        \State $f_{e},\tau=f(\theta_{k}^{i}),\Gamma(\theta_{k}^{i}) $ \Comment{fitness and trajectory}
        \State$f_{i} = \sum_{t=0}^{T-1} \gamma^{T-1-t}\parallel F_{w_f}(\phi_{w_e}(s_{t}),a_{t})-\phi_{w_e}(s_{t+1})\parallel_{2}$
        \Comment{curiosity over trajectory}
        \State $f_{\theta_{k}^{i}}=\varphi(f_{e}-\mu_{f_{e}})/\sigma_{f_{e}}+(1-\varphi)(f_{i}-\mu_{f_{i}})/\sigma_{f_{i}}$
         \Comment{global fitness}
        \State $\mathcal{D}\gets\mathcal{D}\cup\{(s_{j+1},a_j,s_j)\sim U(\tau) | j\in [0,m]\}$
        \Comment{add $m$ transition to $\mathcal{D}$}
    \EndFor
    \State $\{\theta_j | j \in [1,\lambda]\}= sort((\theta_{k}^{i},f_{\theta_{k}^{i}}), i \in [1,\lambda])$ \Comment{sort individuals by fitness}
    \State $\nabla_{\theta_{k}}=\frac{1}{\sigma\mu}\sum_{j=1}^{\mu} (\theta_{j}-\theta_{k})w_{j}$  \Comment{estimate gradient}
    \State $\theta_{k+1}\gets \theta_{k} + \alpha\nabla_{\theta_{k}}$ \Comment{update $\theta$}
    \For{$l=1,p$}
    %\State $L_{I}(w_{i},w_{e})=\parallel I_{w_{i}}(\phi(s_t),\phi(s_{t+1}))-a_t \parallel_2$, $(s_{t+1}, a_t, s_t) \in \mathcal{D}$ \Comment{inverse loss}
    %\State $L_F(w_{f},w_{e})=\parallel F(\phi(s_{t}),a_{t})-\phi(s_{t+1})\parallel_{2}$ , $(s_{t+1}, a_t, s_t) \in \mathcal{D}$ \Comment{forward loss}
    \State $L_{ICM}(w_{i},w_{f},w_{e})=(1-\beta)L_{I}+\beta L_{F}$ \Comment{ICM loss over $\mathcal{D}$}
    \State $w\gets w + \alpha_{ICM} \nabla_{w}L_{ICM}, w \in (w_e, w_i, w_f)$  \Comment{update ICM weights}
    \EndFor
\EndFor
\end{algorithmic}
\end{algorithm}

% shorter explanation with math
In the first generation, we randomly initialize the ES starting center point $\theta_0$, an ICM with its three models, $F_{w_{f}}$, $I_{w_i}$, $\phi_{w_e}$, and an empty replay buffer $\mathcal{D}$.
At each generation, we sample a population of $\lambda$ individuals $\theta_k^{i}$ from a Normal distribution with center $\theta_k$ and standard deviation $\sigma$.
The individuals are then evaluated in the environment and for each individual, the function $f$ returns the extrinsic fitness $f(\theta_{k}^{i})=f_{\theta_{k}^{i}}^{e}$ and the function $\Gamma$ returns the trajectory sampled by the agent in the environment : $\Gamma(\theta_{k}^{i})=\tau=\{(s_{t},a_{t},s_{t+1}), t \in [0,T-1]\}$.
Then, the ICM uses the trajectory $\tau$ to compute the intrinsic fitness $f_{\theta_{k}^{i}}^{i}$ which is equal to the weighted sum of the Curiosity for all transitions in the episode:  
\begin{equation}
  f^{i}=\sum_{t=0}^{T-1} \gamma^{T-1-t}\parallel F_{w_f}(\phi(s_{t}),a_{t})-\phi(s_{t+1})\parallel_{2}
\label{eq:curiosity_f1}
\end{equation}
where $\gamma \in [0,1]$ is a parameter that allows weighting certain parts of the trajectory. If $\gamma\simeq0$, the intrinsic fitness only rewards new transitions explored at the end of an episode, similar to the use of the final state as a behavior descriptor in NS. However, if $\gamma=1$, the intrinsic fitness rewards new transitions seen over the entire trajectory. We found that this was useful when the environment includes multiple obstacles between the beginning states and the final rewarding state, and when the environment has multiple rewarding states. We primarily used $\gamma$ values near 1.

In practice, the intrinsic fitness can be computed at the same time as the normal evaluation of the individual; at each transition, the policy network is used to determine the next action, and once the next state is observed, the ICM is used to calculate Curiosity. As such, the computational overhead of the Curiosity-ES compared to a standard ES is one forward pass through the ICM per transition in the task and $p$ training epochs per generation.

Once the intrinsic fitness of an individual is computed, its total fitness is calculated according to a weighted average of the two normalized finesses:

\begin{equation}
f_{\theta_{k}^{i}}=\varphi(f_{e}-\mu_{f_{e}})/\sigma_{f_{e}}+(1-\varphi)(f_{i}-\mu_{f_{i}})/\sigma_{f_{i}}
\end{equation}
\label{eq:fit}

where $\varphi \in [0,1]$ determines the ratio between extrinsic and intrinsic fitness. In all experiments, we used $\varphi=0.8$ in order to focus mostly on the extrinsic fitness once found. When the whole population has extrinsic fitness values of 0, i.e. reward has not yet been found, we use randomly sampled extrinsic fitness values. We note that we used rank-based fitness, so differences in $\varphi$ only impact the ES update if they change the ranking of individuals, and that, for $\varphi>0.5$ the highest-ranked individual will necessarily be a rewarding policy if any of the policies have found reward. This fitness definition using normalized fitness simplifies the management of hyperparameters by shifting the focus from the magnitude of the two fitness measures to the exploration/exploitation trade-off, represented by $\varphi$, suited to the task.

After calculating the fitness of an individual, some of the transitions are retained in a replay buffer $\mathcal{D}$ in order to train the ICM. To avoid excessive memory usage of the replay buffer, we limit the additions to $m$ transitions per individual, which are sampled uniformly over the trajectory $\tau$. Given unlimited memory, the entire trajectory $\tau$ could be stored in $\mathcal{D}$, however we found that uniform sampling was sufficient for training the ICM.

Once all individuals have been evaluated, they are sorted according to their global fitness. The gradient is then computed using the $\mu$ best individuals and a weight vector $w_j$. We use a weight vector from CMA-ES and other ES \cite{hansen2001completely, chrabaszcz2018back}, $w_j = \frac{log(\mu+0.5)-log(j)}{\sum_{i=1}^\mu log(\mu+0.5)-log(i)}$. We use a learning rate $\alpha$ to control the gradient update to the ES distribution center $\theta_k$.

Finally, we train the ICM networks for $p$ epochs by minimising $L_{ICM}$ with stochastic gradient descent over $\mathcal{D}$. This is equivalent to the ICM training in \cite{pathak2017curiosity}, however the transitions stored in $\mathcal{D}$ are gathered from generations of populations of agents and batches sample transitions from the entire population. We posit that this creates higher diversity in the training set than when using ICM in an RL algorithm, however we noted that the ICM was able to converge quickly over similar areas of the tested environments.

%$f_{\theta_i^k}^{g}=\frac{f_{\theta_{k}^{i}}^{i}+f_{\theta_{k}^{i}}^{e}}{2}$ : $\{r^{j}|j \in [0,\lambda]\}=sort((\theta_{k}^{i},f_{\theta_{k}^{i}}^{g}), i \in [0,\lambda])$.
%The gradient is then computed using the $\mu$ best individuals and $\theta_{k}$ is updated: 
%$$\nabla_{\theta_{k}}\mathop{\mathbb{E}_{\theta_{k}^{i}\sim \mathcal{N}(\theta_{k},\Sigma)}}[F(\theta_{k})]=\frac{1}{\sigma\mu}\sum_{j=0}^{\mu} (r^{j}-\theta_{k})w_{j}$$
%$$\theta_{k+1}\gets \theta_{k} + \alpha\nabla_{\theta_{k}}\mathop{\mathbb{E}_{\theta_{k}^{i}\sim \mathcal{N}(\theta_{k},\Sigma)}}[F(\theta_{k})]$$

Curiosity-ES then continues until termination, here based on a number of generations. As Curiosity is a primarily a modification to the fitness function, more complex termination methods such as population restarts could be used. In this work, we intentionally use a simple ES in order to highlight the capacity of Curiosity to guide search.

\section{Experiments}
\label{sec:experiments}

We use two types of environments to evaluate Curiosity-ES: maze navigation and robotic control. As the primary novelty of Curiosity-ES is the use of Curiosity as intrinsic fitness, we evaluate the utility of Curiosity as an intrinsic motivation in comparison to Novelty, as used by NS-ES \cite{conti2018improving}. To further understand how Curiosity guides evolutionary search, we compare to two state-of-the-art QD methods, CMAME \cite{fontaine2020covariance} and MAP-Elites \cite{cully_robots_2015}, which explicitly search for new behavior. Finally, as we hypothesize that Curiosity could be more beneficial with an ES, we compare our method with a Reinforcement Learning algorithm, specifically TD3 \cite{fujimoto2018addressing}, using ICM to generate exploration bonuses.
% Since, we wanted to compare our method with an approach allowing for a generic definition of the behavior descriptor defined over the all trajectory, we implemented Aurora. We wil put that after i guess

\subsection{Environments}
\label{sec:envs}

\begin{figure}[!h]
    \centering
    % \textbf{Mazes}\par
    \includegraphics[width=0.32\textwidth]{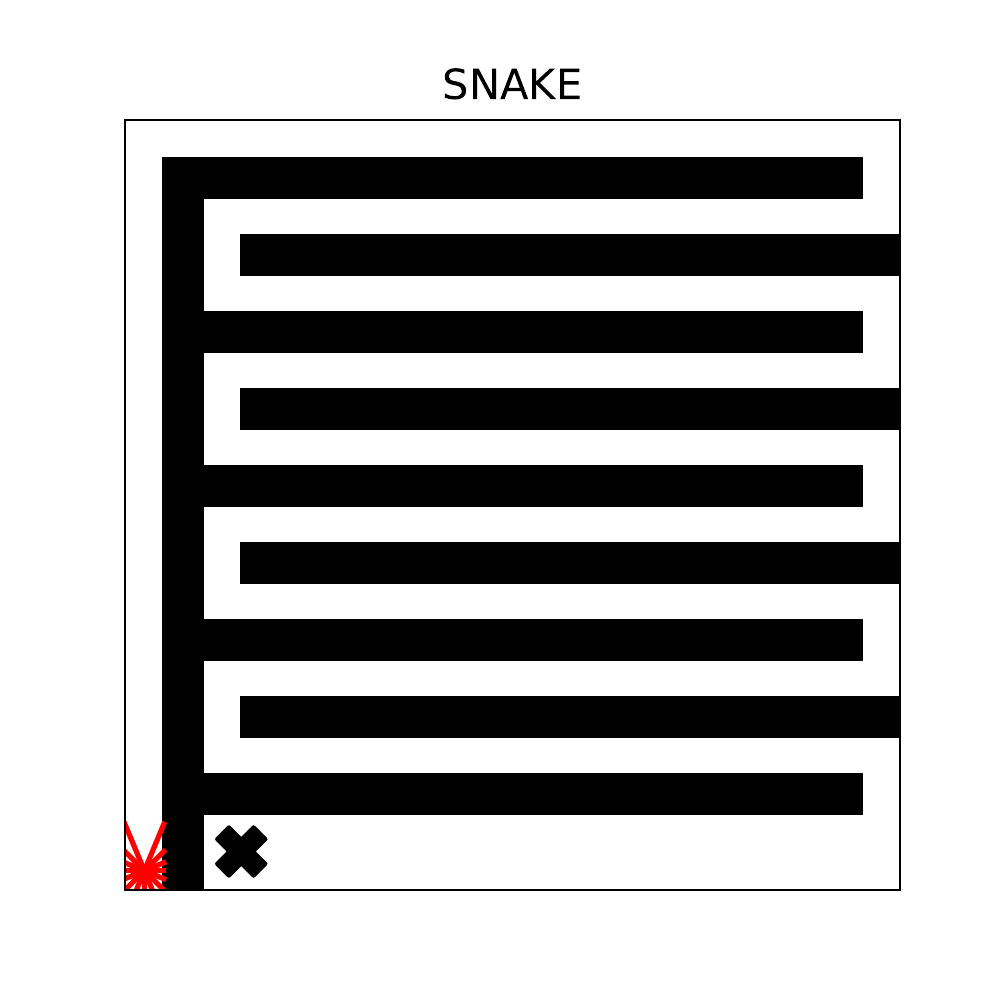}
    \includegraphics[width=0.32\textwidth]{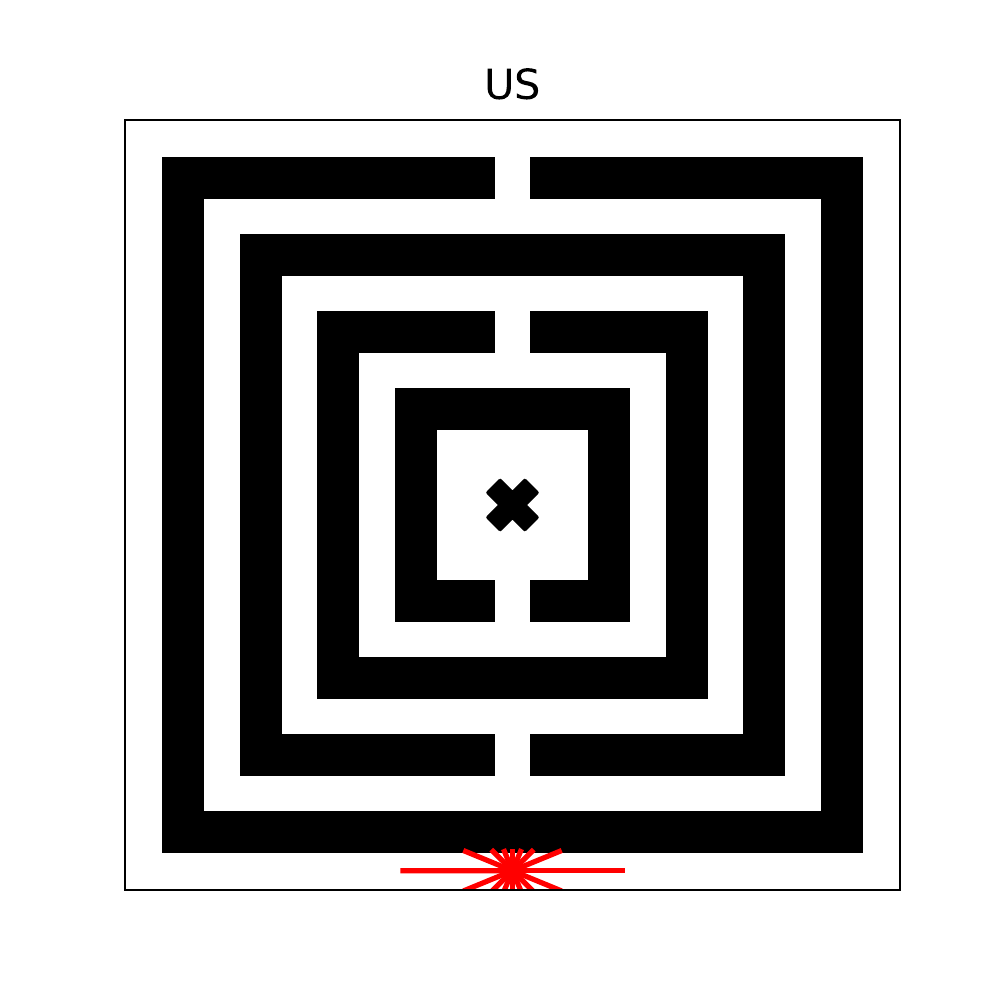}
    \includegraphics[width=0.32\textwidth]{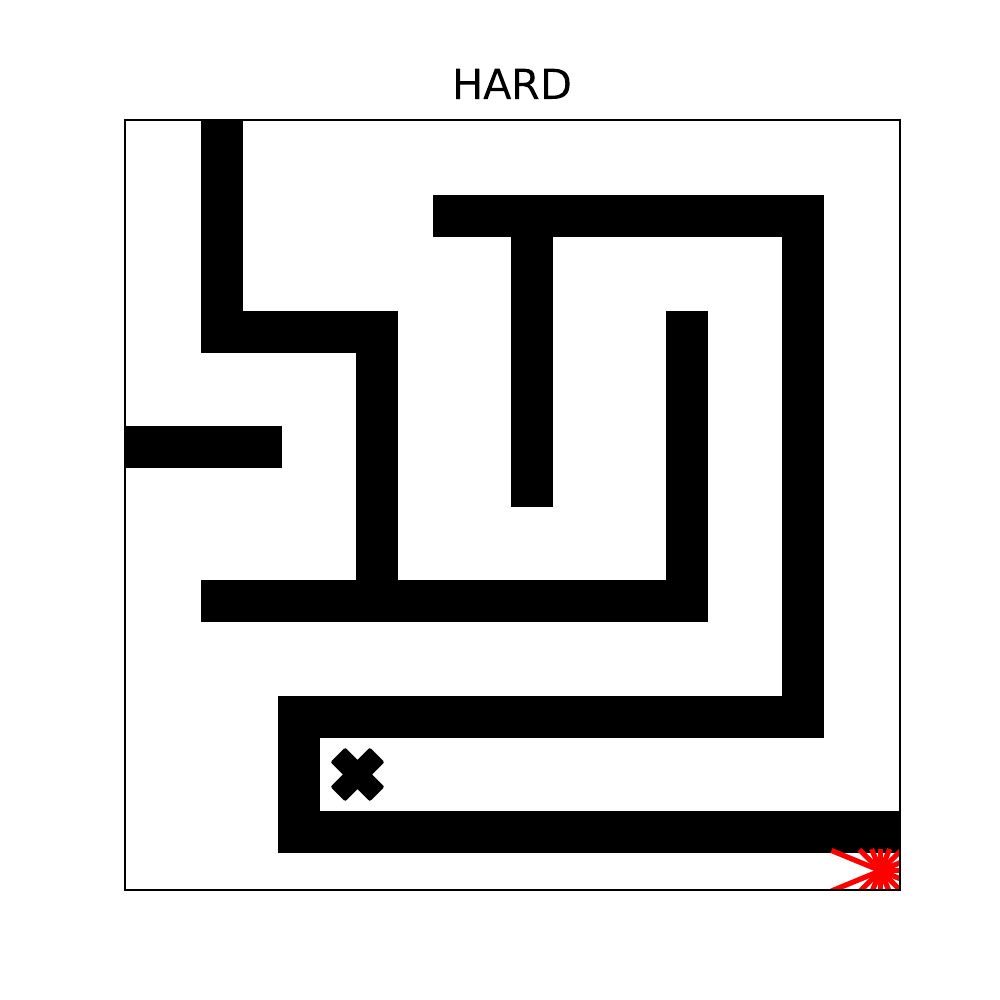}
    \caption{The mazes used for the robotic maze nagivation tasks. Starting point and simulated LIDAR beams are shown in red, with the target marked as X.}
    \label{fig:mazes}
\end{figure}

We use a set of three robot navigation tasks inside mazes of increasing difficulty. Mazes have been commonly used in the study of both Novelty Search \cite{lehman2011abandoning} and sparse rewards in RL \cite{fu2017ex2}. In this work, each maze is a sparse reward MDP where the state space $\mathop{S}$ and the action space $\mathop{A}$ are continuous and agents are rewarded for reaching a unique terminal point efficiently. The mazes used are shown in \autoref{fig:mazes}.

The agent starts in the position $(x_{s},y_{s})$, shown as a red point in \autoref{fig:mazes}, and is rewarded if it reaches the goal $(x_{g},y_{g})$, shown as a black cross. The state space $\mathop{S}$ is the concatenation of the position $(x,y)\in \mathbf{R}^{2}$, the velocity $(v_{x},v_{y})\in \mathbf{R}^{2}$ and a set of simulated LIDAR beams $\{n_i\}_{i \in [0,32]}\in \mathbf{R}^{32}$. These beams return the distance to the nearest wall in 32 directions, allowing the agent to observe obstacles. The policy network outputs an acceleration, defining the movement action of the individual: $\mathop{A}=\{a_x,a_y\}\in\mathbf{R}^{2}$.

The agent is rewarded by $r=(1-\frac{t}{T})$ if it reaches the goal coordinates within a given threshold $\delta$: $\|(x_{t},y_{t}),(x_{g},y_{g})\|_2<\delta$ with $\delta=2$. $T$ is the evaluation time of an individual, so the reward is inversely proportional to the agent's lifetime. This is done to encourage policies which reach the goal state efficiently, however if the goal state is not reached, the reward is simply 0.

To better evaluate Curiosity-ES on different task settings, we also use the DeepMind Control Suite \cite{tassa2018deepmind}, a diverse set of continuous robotic control tasks. Specifically, we use the Ball in Cup, Finger, and Stacker environments. The Ball in Cup catch task requires the agent to move a ball, initially hanging under a cup attached by a string, into a cup; the agent is rewarded 1 only if the ball is caught by the cup. In Finger, a planar `finger' with two joints must rotate a free body on an unactuated hinge; we use the hard task where the tip of the free body must overlap with a small target zone. Finally, we use Stacker, where a gripping robot must stack boxes to a specific target. We use a single box, making this task similar to the Manipulator task also provided in \cite{tassa2018deepmind}. For each of these environments, the agent is only rewarded upon reaching the goal state. As with the navigation environments, we limit the number of possible timesteps, here to $T=1000$, and reward individuals for completing the task more efficiently.

\subsection{Evolutionary Strategy}
\label{sec:es_experiments}

We use Canonical ES \cite{chrabaszcz2018back} as the base for Curiosity-ES and Novelty Search ES in all experiments. The only modification to Canonical ES, beyond adapation to include intrinsic motivation, is the use of a learning rate $\alpha$ on the population center update.

In comparing Curiosity-ES and NS-ES, we vary only the intrinsic motivation in order to isolate the influence of Curiosity on search. This comparison requires adaptation of NS-ES, which we detail here. We compare with NS-ES using Novelty alone and with extrinsic reward, but for simplicity refer to both as NS-ES; in \cite{conti2018improving}, NS with extrinsic reward is termed NRS-ES. We do not maintain a meta-population but rather follow the population method of Canonical-ES, i.e. one population center per generation. We determine the NS-ES hyperparameter of kNN size following the same method as other hyperparameters. For both Curiosity-ES and NS-ES, we set $\varphi=0.8$, meaning that normalized extrinsic fitness contributes 80\% of the final fitness. We use the same behavior descriptor for NS-ES as for the other QD algorithms, detailed below.

All hyperparameters of Canonical ES are shared between Curiosity-ES and NS-ES and vary according to environment; the hyperparameters are presented in \autoref{tab:hyperparameters}. For all the environments we used $\alpha_{icm}=10^{-4}$. Hyperparameters were found after minimal manual tuning and population size parameters were determined by computational node size to maximize parallelization. The number of total evaluations was determined based on convergence and to limit runtime to one day per evolution on a standard CPU cluster. Both methods were implemented using the Ray framework \cite{moritz2018ray} and all environments and hyperparameters have been included in the open-source repository\footnote{\href{https://github.com/SuReLI/Curiosity-ES}{https://github.com/SuReLI/Curiosity-ES}}.

\begin{table}[!h]
\centering
\begin{tabular}{ |p{1.5cm}||p{1cm}|p{1cm}|p{1.2cm}|p{2cm}|p{1.7cm}|p{1.7cm}|}
 \hline
   & SNAKE & US & HARD & BALL IN CUP & FINGER & STACKER \\
 \hline
 $\sigma$   & 0.5 & 0.5 & 0.5 & 0.005 & 0.005 & 0.001\\
 %$\sigma$   & $10^{-1}$  & $5\times10^{-1}$ &   $5\times10^{-3}$ &  $1\times10^{-3}$ &  $5\times10^{-3}$\\
 \hline
 $\mu$     & 56 & 56 & 56 & 56 & 56 & 56\\
 \hline
  $\lambda$     & 28 & 28 & 28 & 28 & 28 & 28\\
 \hline
 $\alpha$   & 0.5 & 1 & 1 & 1 & 0.5 & 0.5\\
 %$\alpha$   & $5\times10^{-1}$  & 1 &   $5\times10^{-1}$ &  1 &  $5\times10^{-1}$\\
 \hline
 \multicolumn{7}{|c|}{NS-ES} \\
 \hline
 $kNN$    & 20 & 10 & 20 & 20 & 10 & 20 \\
 \hline
 \multicolumn{7}{|c|}{Curiosity-ES} \\
 \hline
  $\beta$    & 0.1 & 0.2 & 0.2 & 0.2 & 0.2 & 0.2\\
 %$\beta$    & $10^{-1}$ & $2\times10^{-1}$ & $4\times10^{-1}$ & 2\times10^{-1} & 2\times10^{-1} \\
\hline
$\gamma$    & 0.99 & 0.99 & 0.99 & 0.999 & 0.999 & 0.999\\
%$\gamma$    & 9.9\times10{-1} & 5\times10{-1} & 9.5\times10{-1} & 9\times10{-1} & 9\times10{-1} \\
\hline
 $p$    & 64 & 64 & 64 & 96 & 96 & 64 \\
\hline
\end{tabular}
\caption{Hyperparameters used in the two ES on the six different environments.}
\label{tab:hyperparameters}
\end{table}

\subsection{Comparison with Quality Diversity}
\label{sec:qd_experiments}

While our main goal is to understand how Curiosity functions compared to Novelty as an intrinsic motivation, other QD methods besides NS-ES search explicitly for new behaviors. MAP-Elites, notably, uses mutation from an archive of individuals stored according to their behavior to search for new behaviors. To compare with these QD methods, therefore, it is necessary to define a behavior descriptor suitable for the different tasks and algorithms.

The final state $s_T$ of an individual's behavior has often been used is tasks like maze navigation, locomotion, and robotic control \cite{lehman2011abandoning,paolo2021sparse}. However, in both proposed environments, the state has information not directly related to reward: in the mazes for example, the state contains simulated LIDAR sensor inputs as well as robot velocity. Beyond confounding exploration, this additional information makes the state space intractably large for use as a behavior descriptor for MAP-Elites and CMAME, which require enumerating all possible behavior combinations in a discrete grid. For the behavior descriptor, we therefore use a suitable subset of final states for each task. For mazes, we use the final robot position and velocity. For Ball in Cup, the subset is defined as the position of the ball and the position of the cup. For Finger, the behavior is defined as the two joint positions of the finger plus the joint position of the hinge. Finally, for Stacker, the behavior is defined as the last position of the gripper. In section \ref{sec:bottleneck}, we compare the use of this reward-focused behavior descriptor, which we term $BD_s$, to the use of the full state.

As Curiosity-ES does not require a behavior descriptor, we also compare with the AURORA (AUtonomous
RObots that Realize their Abilities) algorithm \cite{cully_autonomous_2019}, a quality diversity method which uses an autoencoder to learn a behavior latent space. By reconstructing transitions sampled from the individual's entire trajectory, AURORA learns an encoding of transitions which are then used to describe agent behavior. TAXONS \cite{paolo2020unsupervised} is a similar approach which computes a latent representation of the final state; as Curiosity-ES uses the entire trajectory to calculate intrinsic motivation, we compare with AURORA.

We compare Curiosity-ES with five QD algorithms: NS-ES, NSLC, AURORA, MAP-Elites, and CMAME. We use the pyribs \cite{pyribs} library for the implementations of MAP-Elites and CMAME, and include our implementations of NS-ES, NSLC, and AURORA in the provided code. For MAP-Elites and CMAME, we use 50 discrete cells along each dimension of the behavior space for each environments. For MAP-Elites and NSLC we use a Gaussian noise perturbation on the genome as the mutation operator. We use the same mutation standard deviation across all methods. For NSLC, we defined $\mu=7$ and $\lambda=56$, meaning that at each generation of the algorithm the 7 best individual (according to the competition mechanism) produce 8 new individuals each by sampling with Gaussian noise around the expert genome. For CMAME, we used one CMA-ES emitter that acts during phases of 10 generations; we found that this was a good trade-off between exploitation phases and exploration. 

\subsection{Comparing Evolution and Reinforcement Learning}

Finally, in order to understand the benefits of Curiosity for evolutionary search compared to learning, we compare Curiosity-ES to the TD3 algorithm using an ICM for intrinsic motivation, based on \cite{pathak2017curiosity}. While the RL algorithm used in \cite{pathak2017curiosity} was A3C, we believe that TD3 represents a similar and more contemporary choice for the RL algorithm. The implementation of TD3-ICM was based on RLlib \cite{liang2018rllib}.

In order to accurately compare the capacity between the RL method and Curiosity-ES to generate diverse policies, we modify TD3-ICM as follows. At each training step of the algorithm, we sample one roll-out (trajectory) in the environment that is added to the replay buffer; we then train the policy and the ICM using a learning rate of $\alpha_{ICM}=0.6*10^{-5}$. The learning rate for TD3 was set to $\alpha_{TD3}=0.5*10^{-4}$, which we found allowed for a balance of ICM training and TD3 training. Other learning rates quickly led to the ICM learning converging faster than the RL training, making further exploration rely solely on the pseudo-random policy of sampling random actions. 

%and  specific hyper-parameters specified later and we evaluate the policy in the environment. This process is repeated $n=\lambda*N$, $\lambda$ being the number of individuals for the population based methods and $N$ the number of generations.

%The fine tuning of the reinforcement learning algorithm was harder. Indeed, in practice, we saw that ICM was learning the environment much faster than the Reinforcement Learning agent ending in a situation where the exploration rely on the pseudo exploration. We tried to increase the learning rate of the agent in order to find the equilibrium that we defined earlier but it brought huge instability to the TD3 algorithm which in itself is a hard mechanism to fine-tune. Therefore the more stable setting that we found for those environment is to set the learning rate of the ICM to 

For all comparisons, both with other QD methods and TD3-ICM, we performed manual hyperparameter tuning and the algorithm modifications as detailed above in order to improve their performance and present a fair comparison. As seen in the next section, Curiosity-ES finds more efficient policies and covers more of the behavior space than all other tested methods.

\section{Maximising reward and state space coverage}
\label{sec:reward}

We first present a comparison on three maze navigation tasks of a subset of the described methods: Curiosity-ES, NS-ES, MAP-Elites, CMAME, and TD3-ICM. This subset was chosen as these methods were able to find reward in at least one of the six environments. We then evaluate this same set of algorithms on the DMCS tasks Ball in Cup, Finger, and Stacker. Finally, in \autoref{sec:bottleneck}, we present the results of NS-ES, NSLC, and AURORA in a study on the impact of behavior descriptors. We note that, while reward is only given at the end of the episode and is therefore sparse, the amount of reward depends on the efficiency of the policy; Curiosity-ES not only finds reward on all six environments, but also finds competitive or more efficient policies than other reward-finding methods.

%compare the performance of Curiosity-ES, NS-ES, MAP-Elites, CMAME, and TD3-ICM in their capacity to bring policy search to states which give reward in sparse environments. To this end, we evaluate all the fitness based algorithms using a fitness function $f$ which heavily weights reward once it is found, \autoref{eq:fit} with $\varphi=0.8$. Both environments reward efficiently finding the objective, so once evolution has found a rewarding policy, we expect to see continued improvement towards more efficient policies. In the following part we will specifically address the comparison between Curiosity-ES, NS-ES, MAP-Elites, CMAME and TD3-ICM on the specified benchmarks. In the next part, we will go through the experiments including NSLC, $\text{NSLC\_BD}_{s}$, $\text{NS-ES\_BD}_{s}$ and AURORA.

\subsection{Maze Navigation}
\label{sec:maze_navigation}

\begin{figure}[!h]
    \centering
    \textbf{Reward on Maze Navigation}\\
    \raisebox{1.15cm}{\includegraphics[width=0.10\textwidth]{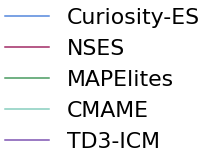}}
    \includegraphics[width=0.29\textwidth]{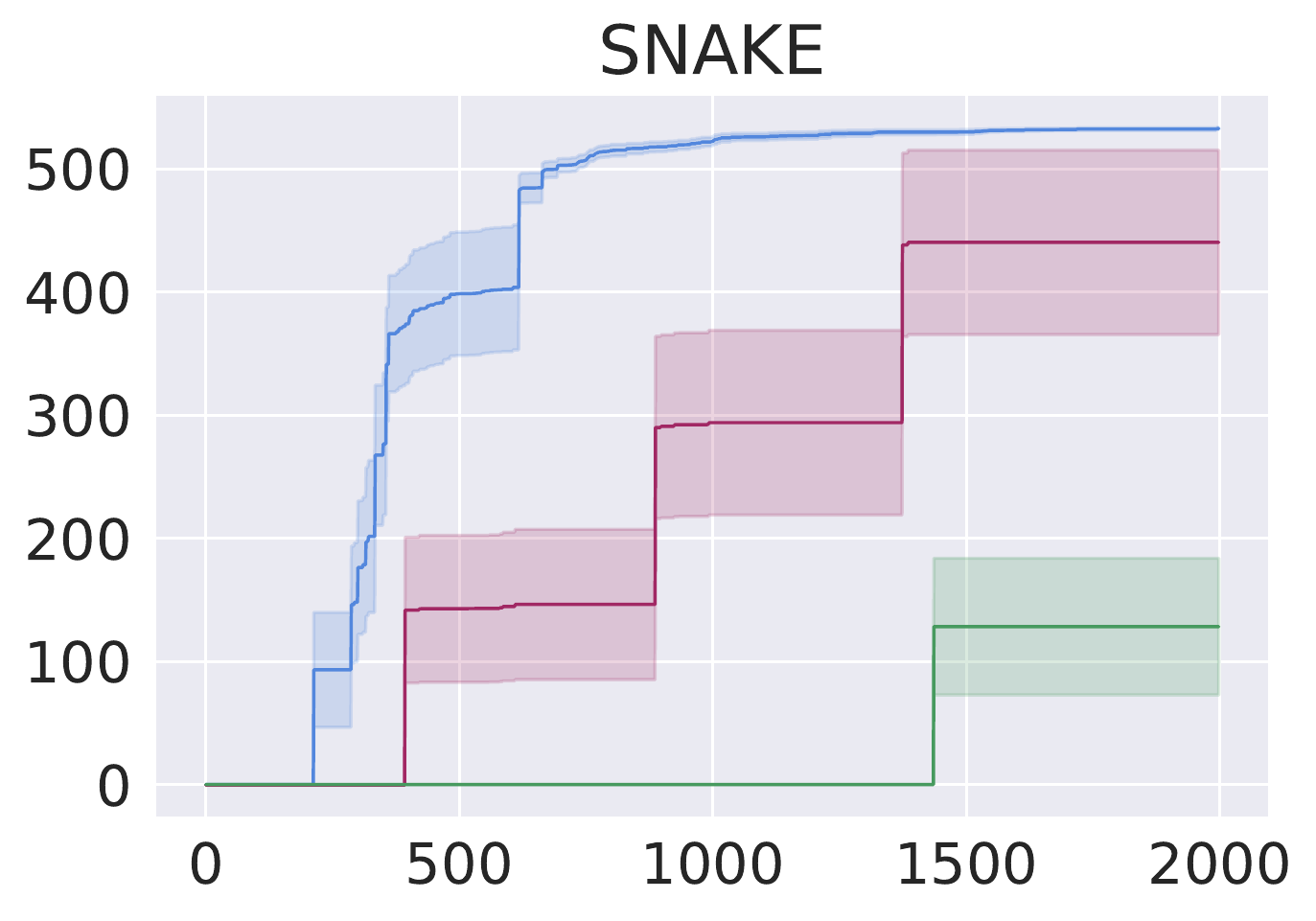}
    \includegraphics[width=0.29\textwidth]{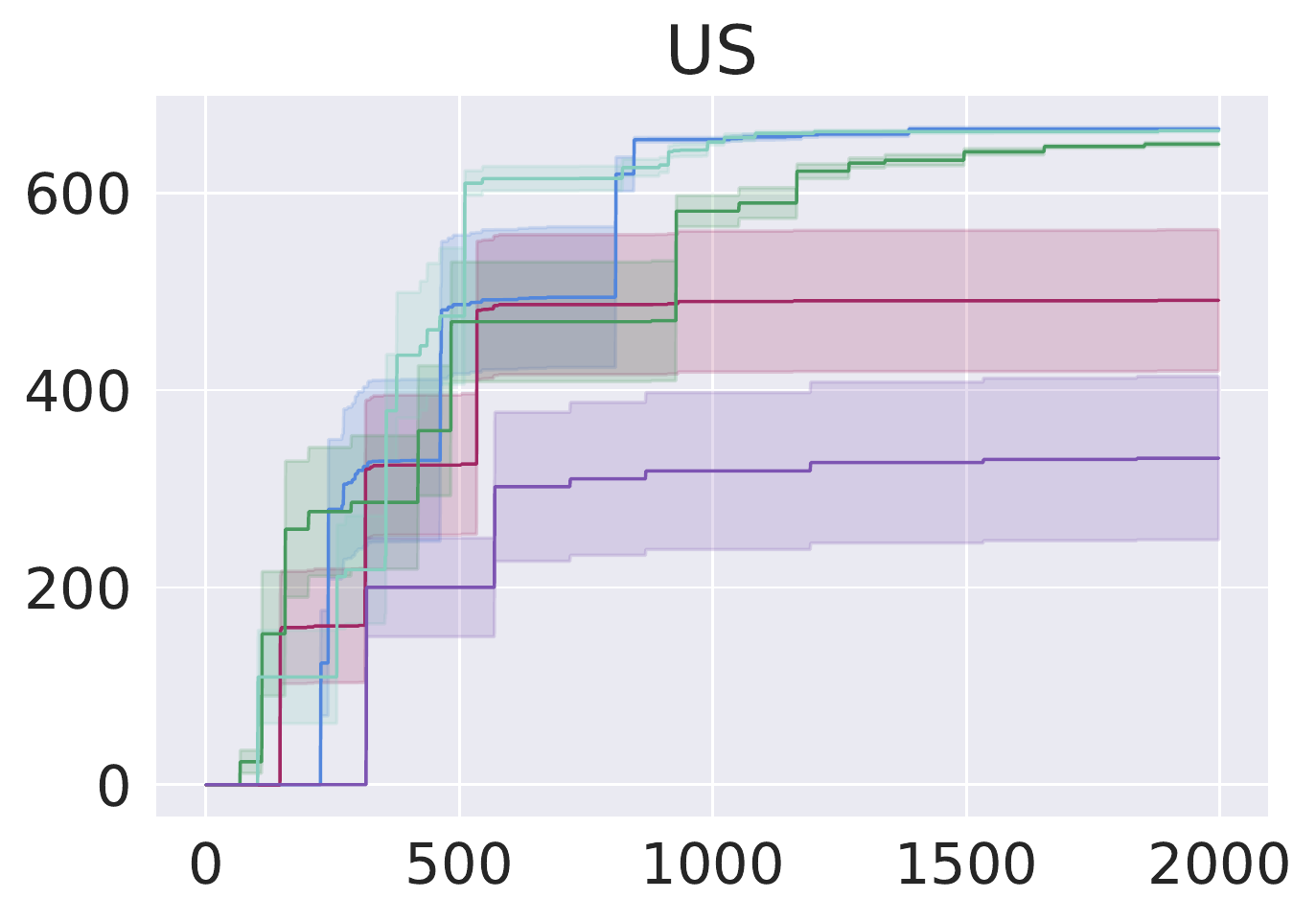}
    \includegraphics[width=0.29\textwidth]{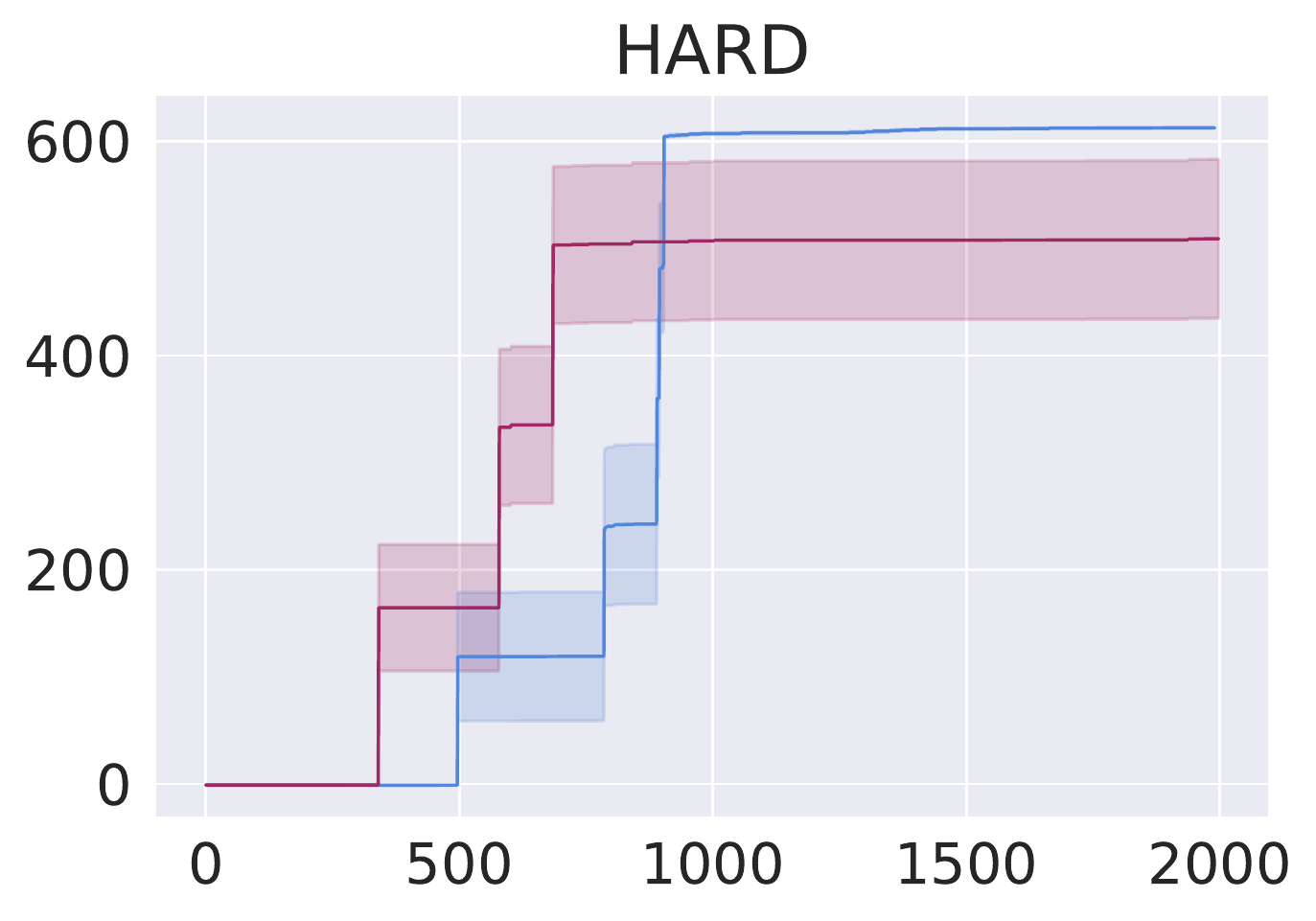}
    \caption{The maximum fitness sampled since the beginning of optimization on the three maze environments. Algorithms which did not find the reward in the environment are not displayed.}
    \label{fig:maze_reward}
\end{figure}

We first study the ability of the selected methods to find rewarding policies on the proposed maze navigation tasks, presented in \autoref{fig:maze_reward}. We note that Novelty Search has been frequently studied on maze tasks and shown to be highly effective for this domain \cite{lehman_exploiting_2008,lehman2011abandoning,mouret2011novelty}. We observe that NS-ES does perform well on all maze environments, leading to policies which find the reward early in search.  However, we note that the found policies often reach an efficiency plateau and do not continuously improve, despite the efficiency reward bonus. Curiosity-ES, on the other hand, continuously improves on rewarding policies by finding more efficient policies throughout search; we believe that this is due to the pressure from Curiosity to continue exploring the policy space beyond new final positions. We propose a further analysis of this specific mechanism in \autoref{sec:diversity_rewarding_policy}.

Additionally, we note that CMAME only finds rewarding policy on the US maze, and MAP-Elites does not find any such policies on the HARD maze. However, surprisingly, we note that both the two methods outperformed NS-ES for the US environment. We believe that the intrinsic exploration mechanism of MAP-Elites, mutation from experts which reached different final positions, is able to effectively exploit the symmetry of the US maze. CMAME improves on MAP-Elites in this environment using CMA-ES for mutations, leading to efficient policies early in the search. We note that Curiosity-ES also arrives at this level of policy efficiency despite using a simpler ES which does not calculate genome covariance.

Finally, we note that TD3-ICM is also only able to find a rewarding policy on the US maze. Even with manual learning rate tuning, we found that the ICM was able to learn much faster than the TD3 agent explored. As previously mentioned, this leaves TD3-ICM functioning similarly to base TD3, as the intrinsic motivation is small and constant. Even in the US maze, we noted high oscillation in rewarding policies; while \autoref{fig:maze_reward} shows the maximum reward since the beginning of optimization, the learning process of TD3-ICM diverged after finding rewarding policies and was unable to recover. We believe that this is due to variability in the replay buffer, which, in a sparse setting, can be mostly full of non-rewarding transitions.

%has strong difficulty to explore efficiently and find the reward. In practice, we saw that even with our setting, the curiosity module was able to learn much faster the dynamic of the environment than TD3 the intrinsic reward, leading to sub exploration mechanism in some cases. Even on US where the agent was able to found the reward we can see that the algorithm wasn't able to fully focus on the reward in order to optimize the policy. Indeed, in most of the case, the algorithm was able to found only a few rewarding policies but once lost the policy was unable to recover it. 

%In RL, the mechanism of learning the policy highly rely on the batch of data given during the optimization step. In sparse reward setting this implies that transitions leading to the reward will be extremely poor and sampled with low probability. While in ES, the center of the distribution will probably end up on the rewarding policy, allowing to sample in the rewarding policies distribution.

\begin{figure}[!h]
    \centering
    \textbf{Coverage on Maze Navigation}\\
    \raisebox{1.15cm}{\includegraphics[width=0.1\textwidth]{new_figures/principal_legend_final.png}}
    \includegraphics[width=0.29\textwidth]{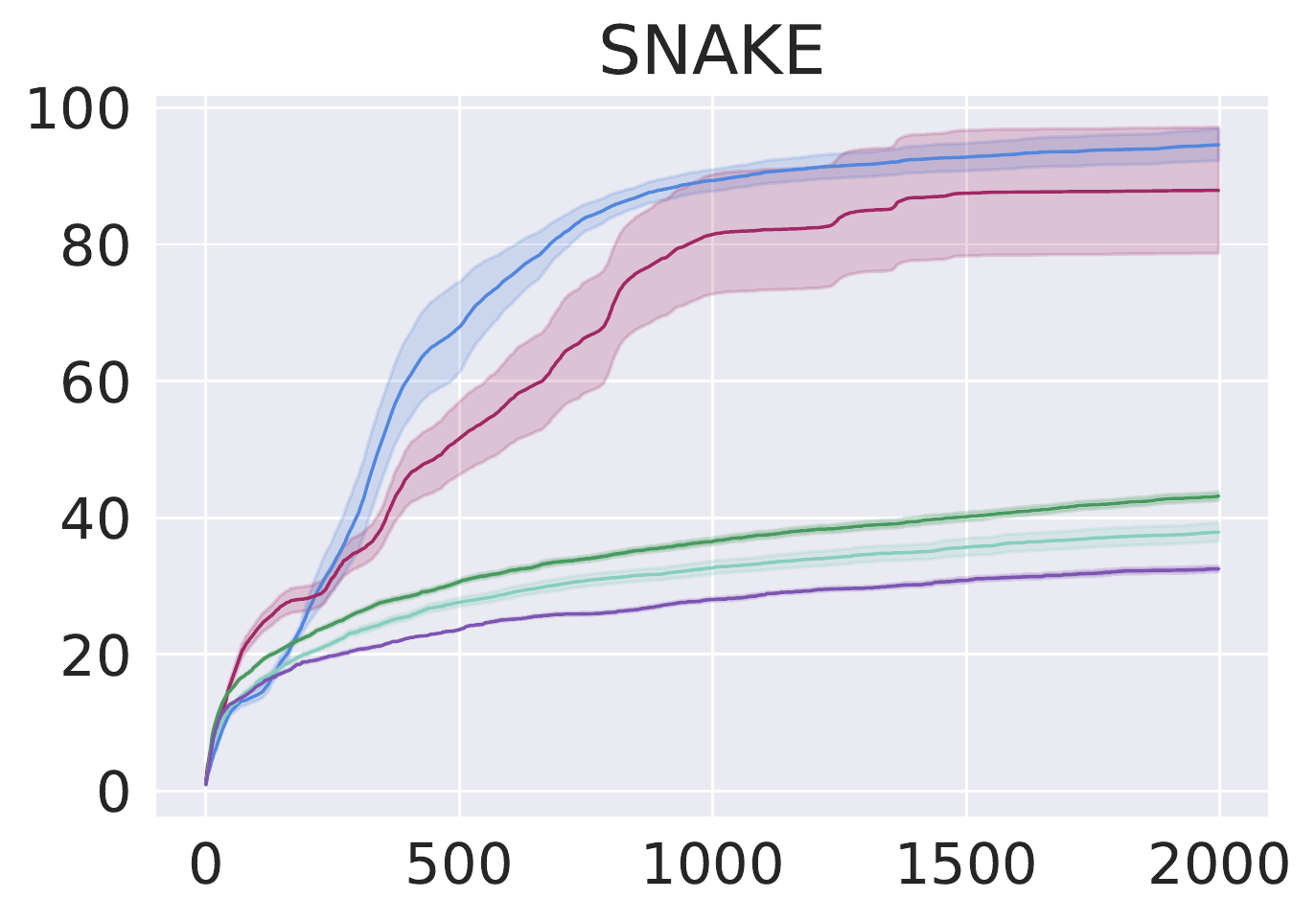}
    \includegraphics[width=0.29\textwidth]{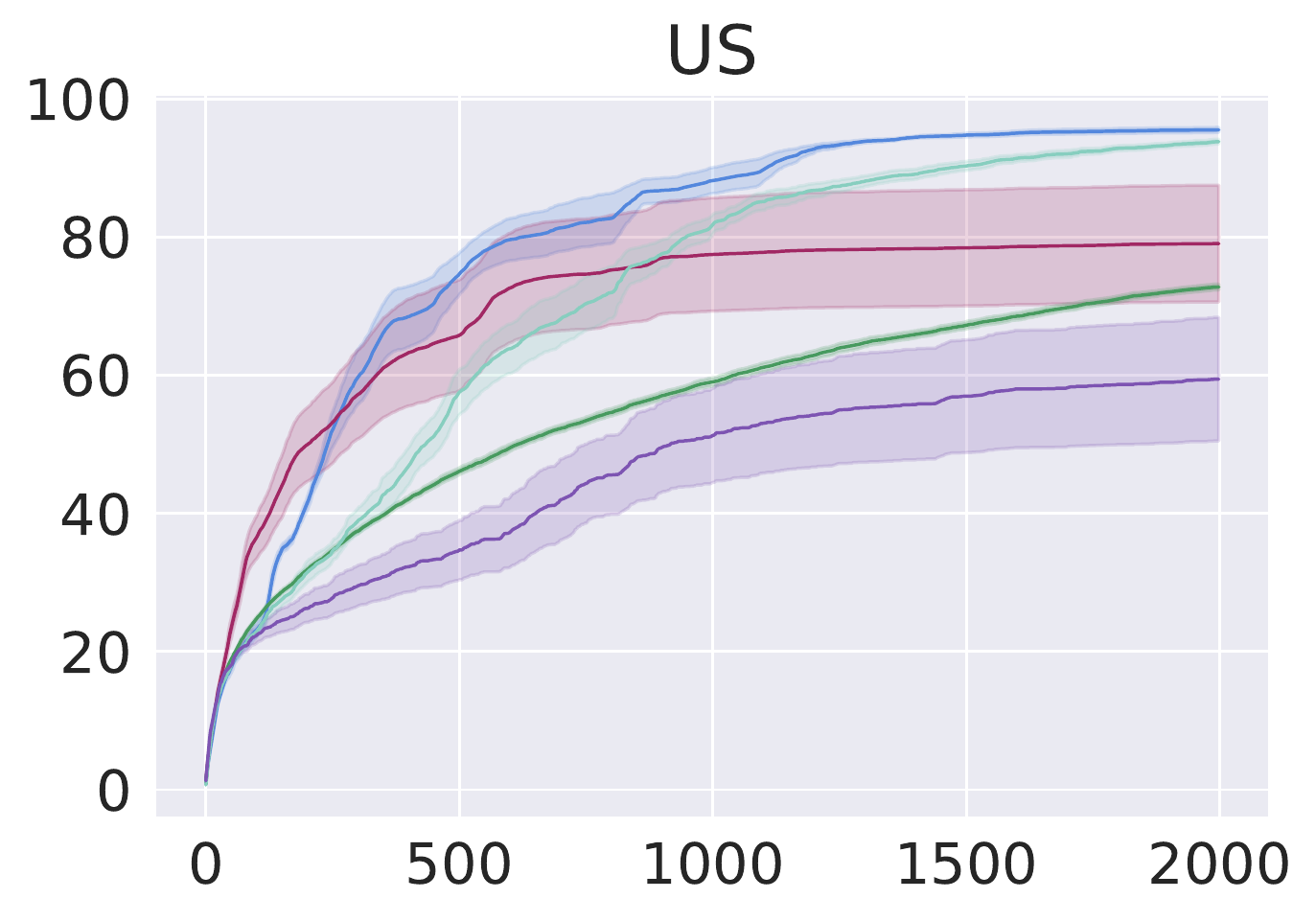}
    \includegraphics[width=0.29\textwidth]{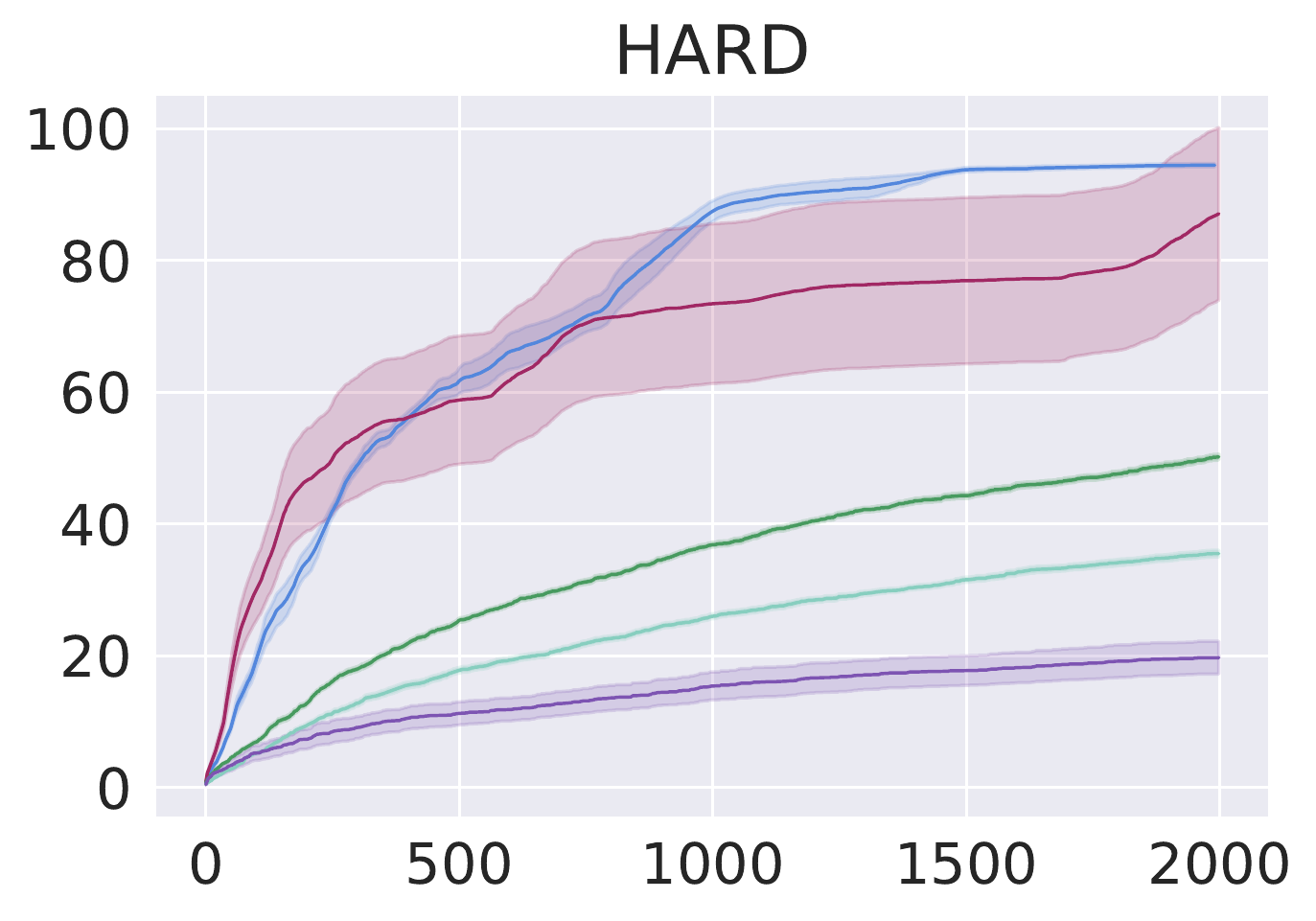}
    \caption{The percentage of discrete final robot positions reached throughout policy search.}
    \label{fig:maze_coverage}
\end{figure}

In \autoref{fig:maze_coverage}, we show the percentage of discretized final robot positions found by each method throughout search. We note that Curiosity-ES continually explores throughout evolution, reaching almost all possible terminal states despite using the entire trajectory for the computation of Curiosity. We also note that, while MAP-Elites and CMAME search explicitly to increase this behavior representation, only CMAME on the US maze is able to explore as effectively as Curiosity-ES.

\begin{figure}[!h]
    \centering
     \begin{subfigure}[b]{0.27\textwidth}
         \centering
         \includegraphics[trim={1cm 1cm 1cm 1cm}, clip, width=\textwidth]{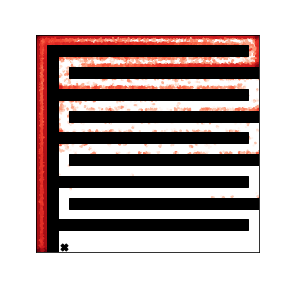}
         \caption{CMAME on SNAKE}
     \end{subfigure}
     \hspace*{\fill}
     \begin{subfigure}[b]{0.27\textwidth}
         \centering
         \includegraphics[trim={1cm 1cm 1cm 1cm}, clip, width=\textwidth]{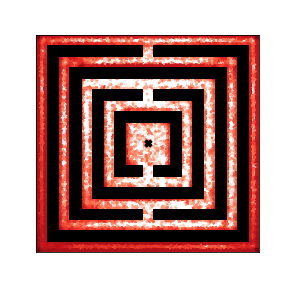}
         \caption{CMAME on US}
     \end{subfigure}
     \hspace*{\fill}
     \begin{subfigure}[b]{0.27\textwidth}
         \centering
         \includegraphics[trim={1cm 1cm 1cm 1cm}, clip, width=\textwidth]{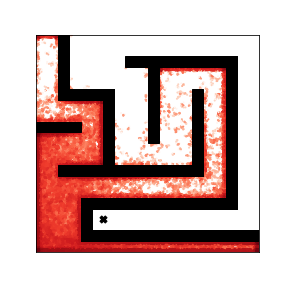}
         \caption{CMAME on HARD}
     \end{subfigure} \begin{subfigure}[b]{0.27\textwidth}
         \centering
         \includegraphics[trim={1cm 1cm 1cm 1cm}, clip, width=\textwidth]{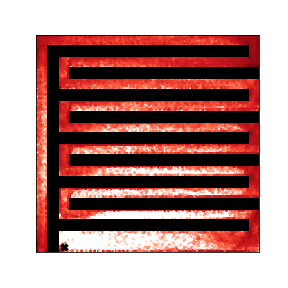}
         \caption{NS-ES on SNAKE}
     \end{subfigure}
     \hspace*{\fill}
     \begin{subfigure}[b]{0.27\textwidth}
         \centering
         \includegraphics[trim={1cm 1cm 1cm 1cm}, clip, width=\textwidth]{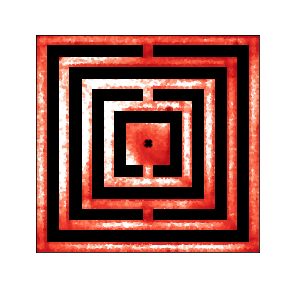}
         \caption{NS-ES on US}
     \end{subfigure}
     \hspace*{\fill}
     \begin{subfigure}[b]{0.27\textwidth}
         \centering
         \includegraphics[trim={1cm 1cm 1cm 1cm}, clip, width=\textwidth]{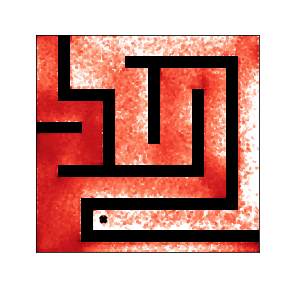}
         \caption{NS-ES on HARD}
     \end{subfigure}    \begin{subfigure}[b]{0.27\textwidth}
         \centering
         \includegraphics[trim={1cm 1cm 1cm 1cm}, clip, width=\textwidth]{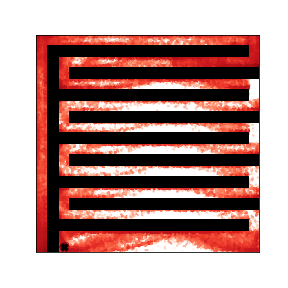}
         \caption{Curiosity-ES on SNAKE}
     \end{subfigure}
     \hspace*{\fill}
     \begin{subfigure}[b]{0.27\textwidth}
         \centering
         \includegraphics[trim={1cm 1cm 1cm 1cm}, clip, width=\textwidth]{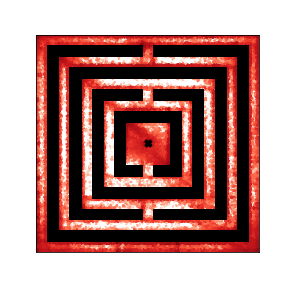}
         \caption{Curiosity-ES on US}
     \end{subfigure}
     \hspace*{\fill}
     \begin{subfigure}[b]{0.27\textwidth}
         \centering
         \includegraphics[trim={1cm 1cm 1cm 1cm}, clip, width=\textwidth]{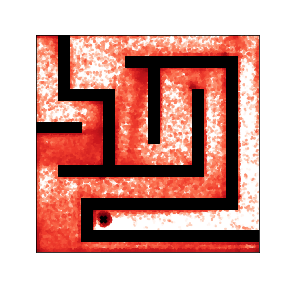}
         \caption{Curiosity-ES on HARD}
     \end{subfigure}
    \caption{Final states reached in the Maze Navigation task by CMAME (top), NS-ES (middle), and Curiosity-ES (bottom). Individual points are semi-transparent; color indicates density.}
    \label{fig:maze_final_states}
\end{figure}

In \autoref{fig:maze_final_states}, we show the final position of each policy during evolution, using a demonstrative evolution for each image. We study CMAME, NS-ES, and Curiosity-ES as NS-ES and Curiosity-ES consistently found reward and CMAME achieved similar coverage and efficiency on the US maze. We first note that, of the three methods, NS-ES has the most even distribution of final states throughout the mazes as Novelty is based on the Euclidean distance of the final position. The high density of similar individuals in NS-ES is influenced by the hyper-parameter $k$, the number of neighbors used in the comparison, which in our case was 10. CMAME has a highly uniform distribution on the US maze and on the explored sections of the SNAKE and HARD mazes, however the low number of policies which reach terminal states between the starting point and the reward leads to a lack of exploration near the objective.

The terminal states of Curiosity-ES demonstrate the utility of Curiosity to find areas which are different from other areas. For example, there are few individuals which terminate in the middle of the SNAKE maze, due to the similarity of transitions in this area. On US, also, self-similar areas in the corridors of the maze are less explored. We argue that this is more fitting for the exploration of regions which are sufficiently different from previously seen areas: in other words, areas which have high Curiosity. On all three mazes, we note that Curiosity-ES arrives at a higher concentration of terminal states leading to the objective point. This exploration around the objective enables the search for more efficient trajectories seen in \autoref{fig:maze_reward}.

The difference in results between US, which has multiple possible rewarding trajectories, and the SNAKE and HARD mazes, which are much more constrained, leads us to the following conclusion. In unconstrained environments such as US, the mutation mechanism of MAP-Elites is effective as a uniform exploration of the environment can more easily lead to new states and the objective. However, when the environment is constrained as in SNAKE and HARD, an informed mutation coupled with an intrinsic motivation that rewards exploration is necessary. While Novelty, as an intrinsic motivation, leads to consistent coverage of many terminal states due to the behavior descriptor of final position, Curiosity naturally leads to an exploration of trajectories, and therefore final states, which cover novel areas of the maze.

%sparse reward environments, when trajectories are highly constrained, an informative mutation coupled with the appropriate intrinsic motivation is much more efficient in finding new policies leading to new area of the environment while for unconstrained environments the simple mechanism of MAP-Elites is already effective.In order to accurately assess the insight of each algorithm in there way of behaving during the exploration, we show in \autoref{fig:maze_final_states} the last state sampled for all the individuals evaluated in each maze.

%Finally, we can see that for the two highly constrained mazes SNAKE and HARD, CMAME struggle in exploring the state space while in US the exploration is much more uniform. The evaluation of this algorithm puts even more in advance the mechanism of exploration of ES by intrinsic motivation. Indeed, we can see that NS-ES and Curiosity-ES follow very specific trends, exploration lines. Where CMAME explores the environment in a much more uniform way (no field lines can be identified).

%In this part we saw the capacity of Curiosity-ES in exploring a set of different mazes. In the next part we will see how the algorithm behave in more heterogeneous sparse environments.

\subsection{Control Tasks}
\label{sec:control_tasks}

We next evaluate the various methods on the sparse DeepMind Control Suite environments using the fitness function defined in \autoref{eq:fit}. These environments reward manipulating the state of the environment, such as rotating an object, rather than moving the agent. Here we define the behavior descriptor as being a subset of the last sampled for each individual.

\begin{figure}[!h]
    \centering
    \textbf{Reward on Control Tasks}\\
     \raisebox{1.15cm}{\includegraphics[width=0.10\textwidth]{new_figures/principal_legend_final.png}}
    \includegraphics[width=0.29\textwidth]{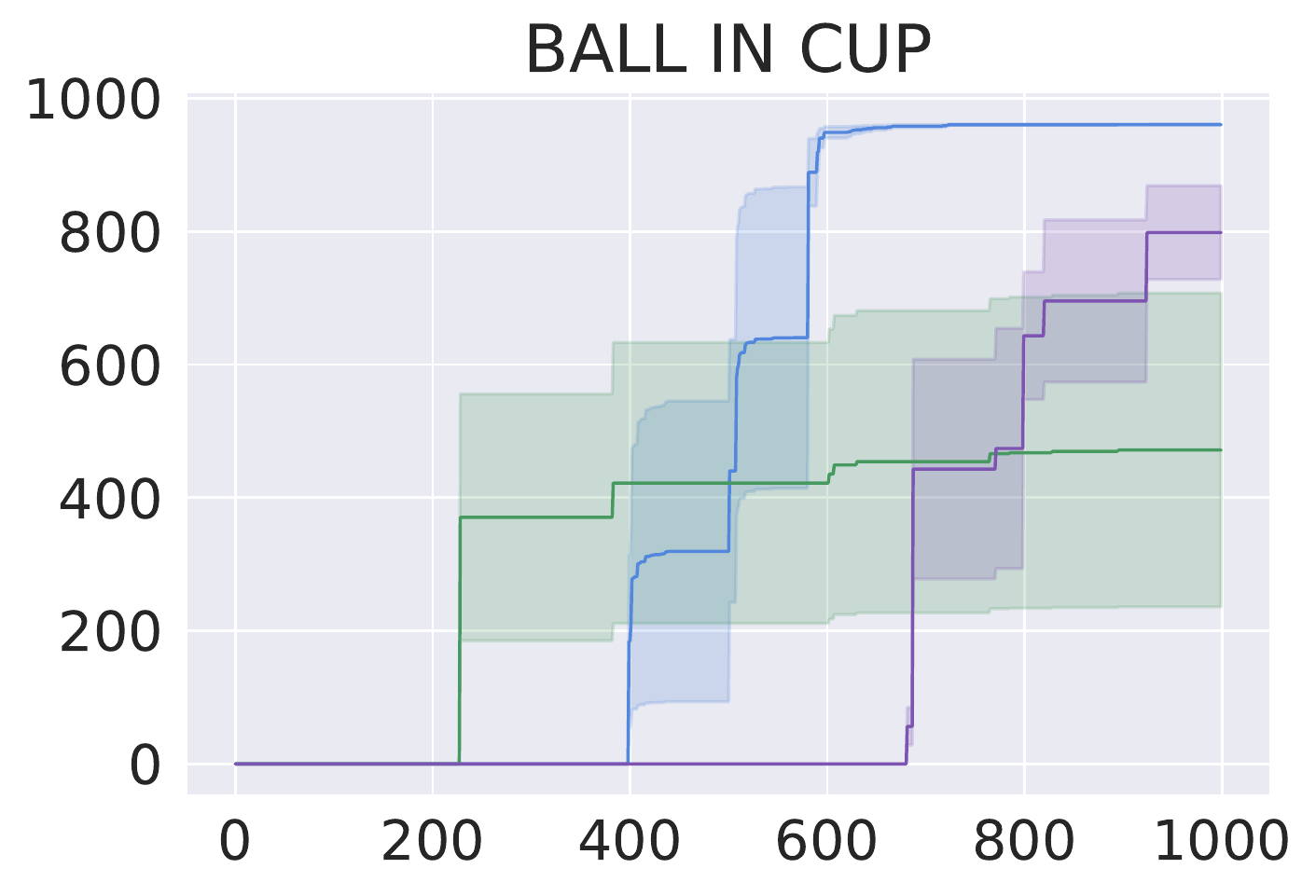}
    \includegraphics[width=0.29\textwidth]{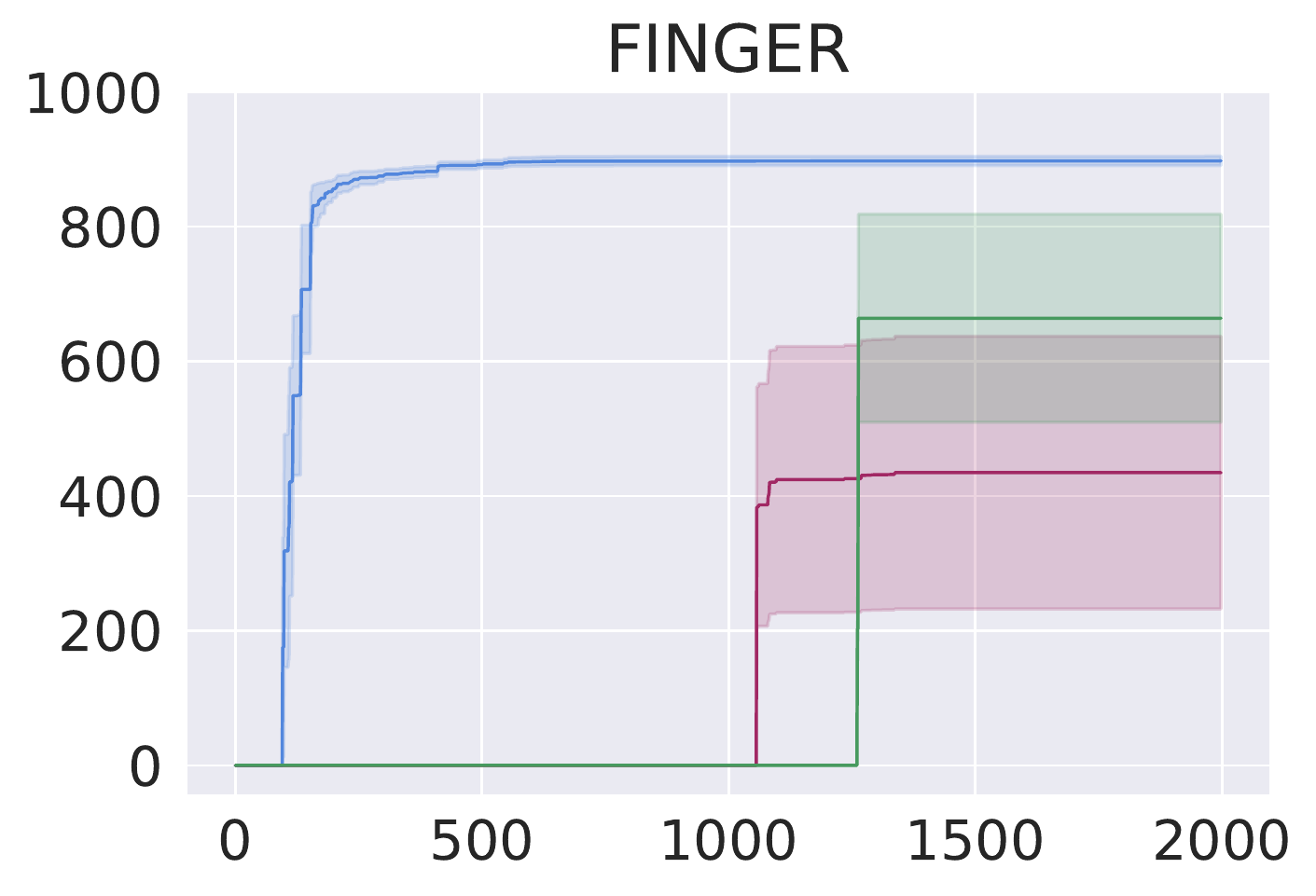}
    \includegraphics[width=0.29\textwidth]{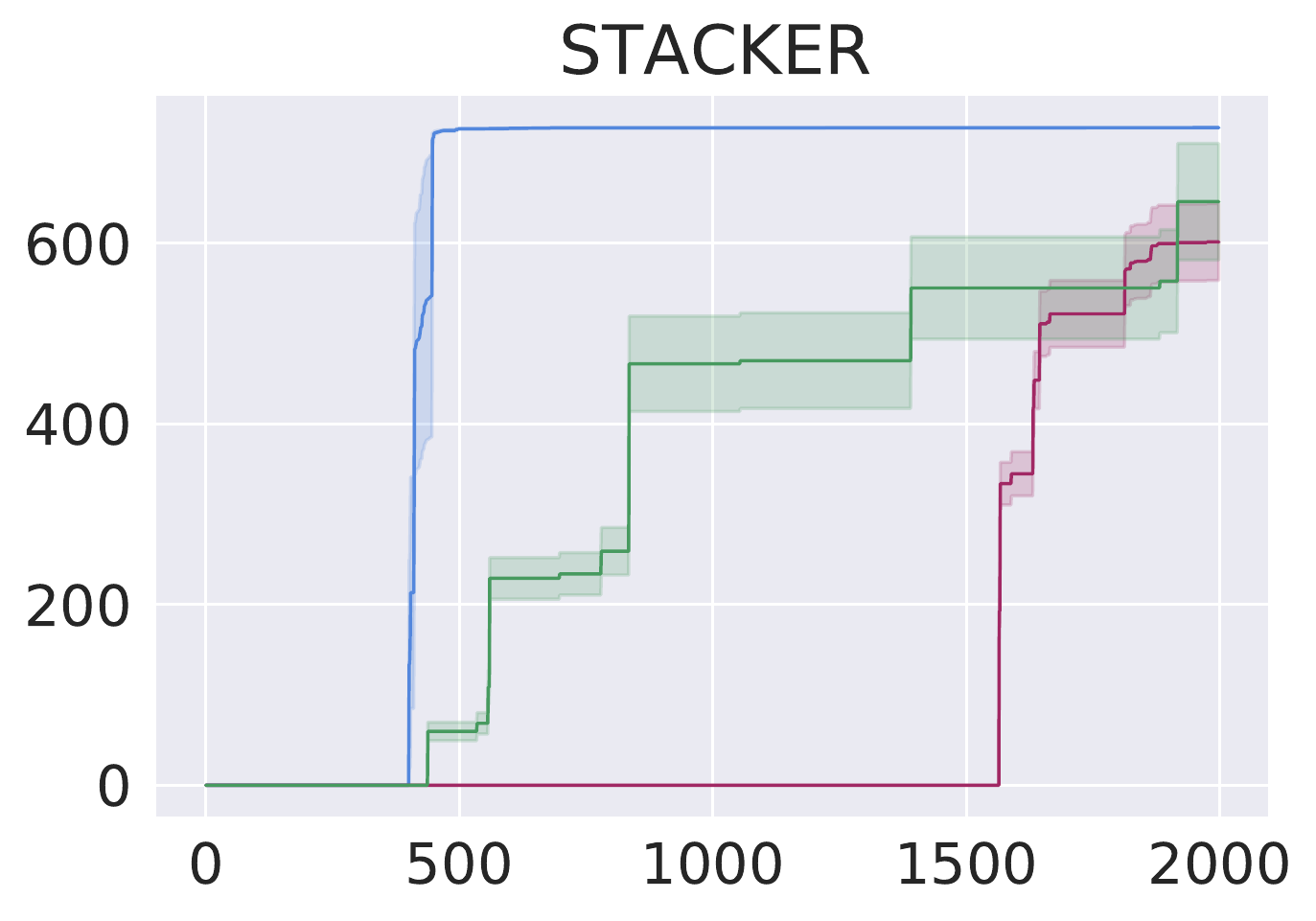}
    \caption{The sum reward of the best policy in each generation on the three control task environments. Algorithms which did not find reward, such as CMAME, are not displayed.}
    \label{fig:control_reward}
\end{figure}

We compare the ability of the different methods to find rewarding policies in \autoref{fig:control_reward}. We first note that Curiosity-ES was able to consistently find efficient rewarding policies on all three environments. MAP-Elites was also able to find rewarding policies on all three tasks, however CMAME did not find rewarding policies on any of the tasks. We found that MAP-Elites lead to better exploration on these tasks than CMAME, as further discussed below. NS-ES was able to find rewarding policies on the Stacker and Finger environments, however on the Finger environment it quickly diverged and did not improve on the efficiency of previous rewarding policies. TD3-ICM was only able to find reward on the Ball in Cup task, however it was able to continuously improve the found policy to increase efficiency.

%The figure \autoref{fig:control_reward} shows once again that TD3-ICM is able to find the reward only in one environment. However it does seem that in this specific setting TD3 was able to induce a reinforcement mechanism were we can see a maximisation of the reward once reached. NS-ES was able to find the rewarding state on two environments. However we can see that the reinforcement mechanism only occurred in STACKER while the optimisation process was highly unstable in FINGER. Additionally, it's interesting to see that MAP-Elites find the reward in the three environment and even outperform NS-ES on FINGER. Curiosity-ES once again find the rewarding state in all the environments. The ES used with the curiosity mechanism is able to focus more on the rewarding state while optimizing the trajectory, leading to more rewarding policies.

\begin{figure}[!h]
    \centering
    \textbf{Coverage in Control Tasks}\\
     \raisebox{1.15cm}{\includegraphics[width=0.10\textwidth]{new_figures/principal_legend_final.png}}
    \includegraphics[width=0.29\textwidth]{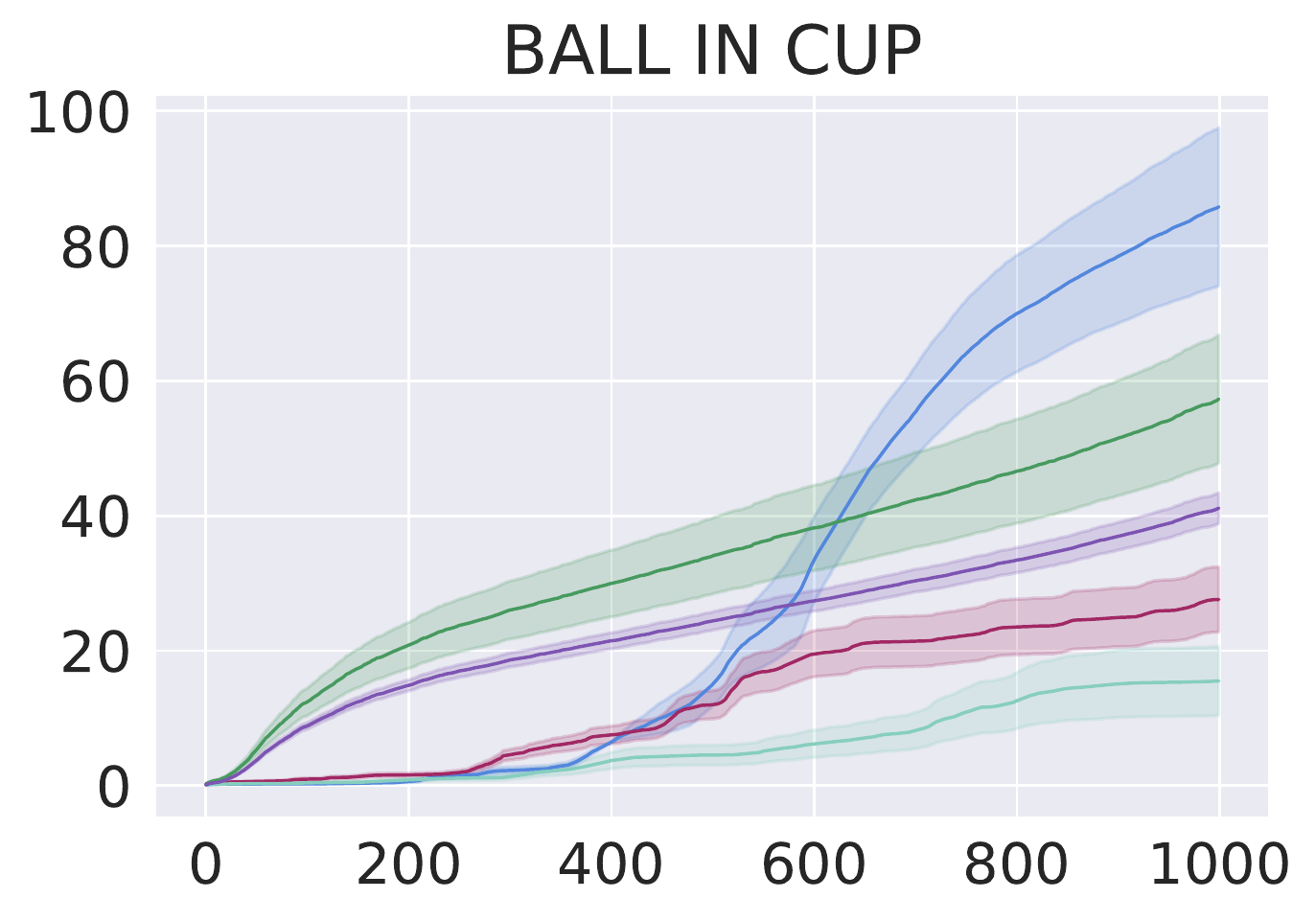}
    \includegraphics[width=0.29\textwidth]{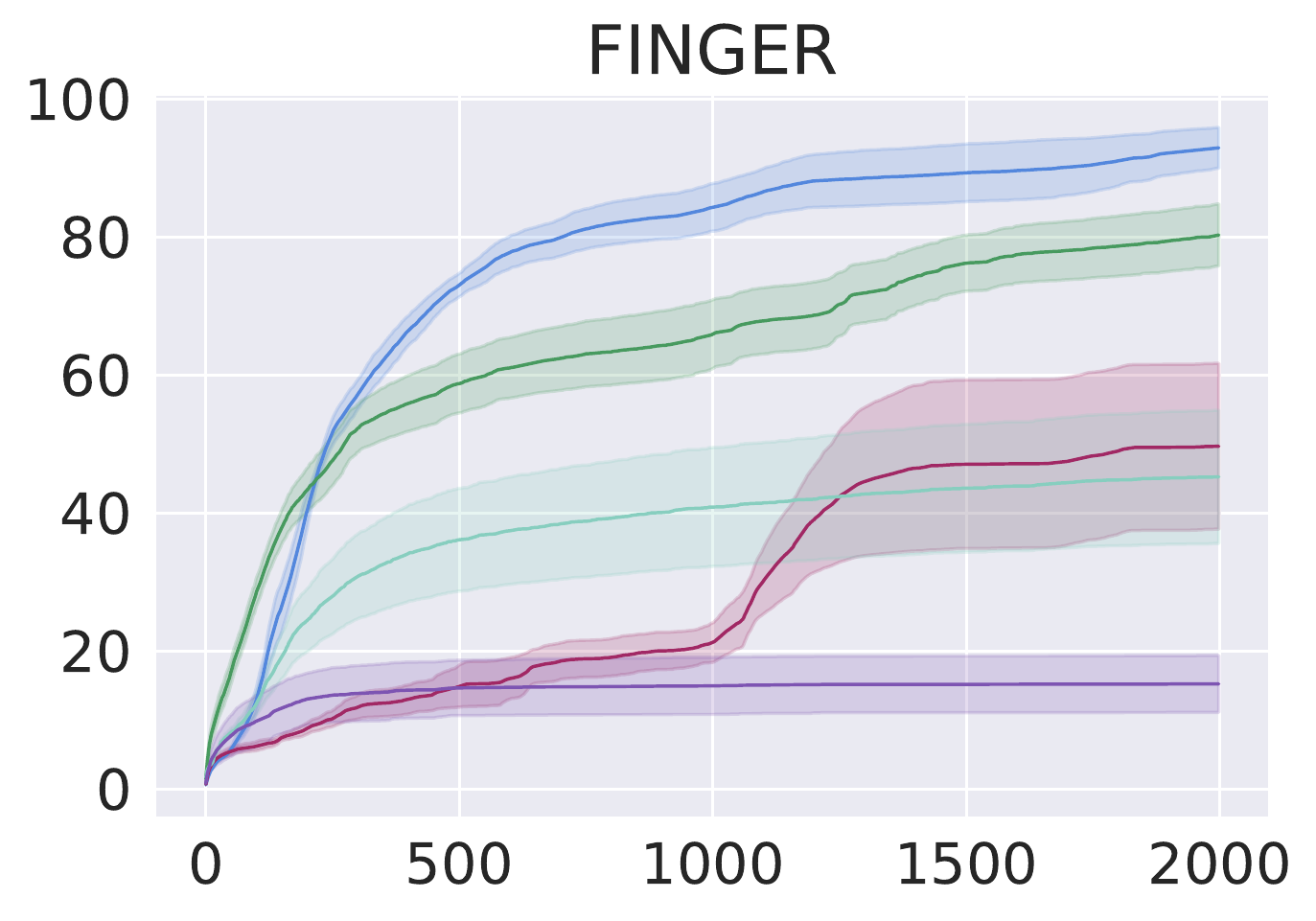}
    \includegraphics[width=0.29\textwidth]{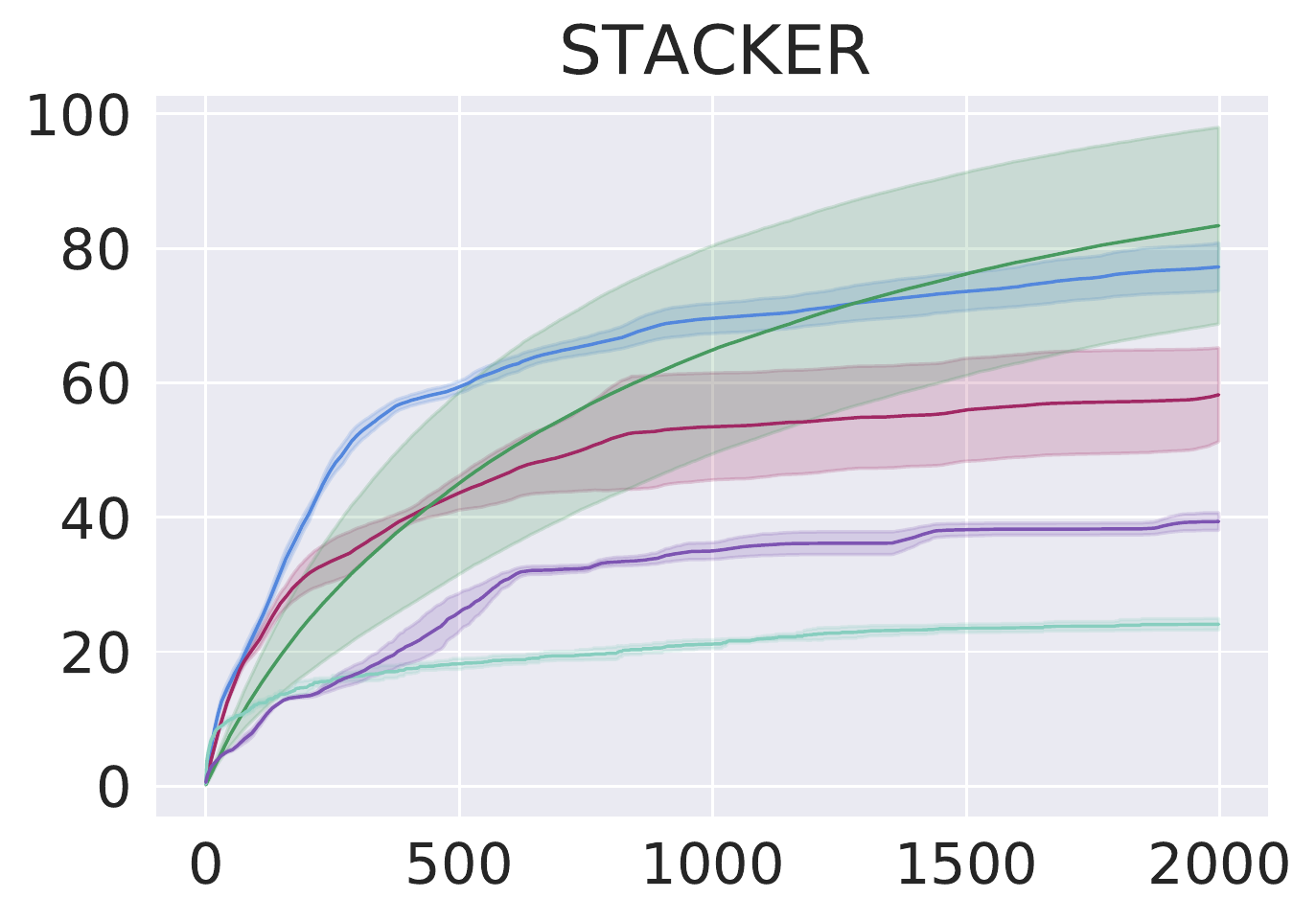}
    \caption{The percentage of discrete terminal behaviors reached throughout policy search.}
    \label{fig:control_coverage}
\end{figure}

% coverage
In \autoref{fig:control_coverage}, we see the coverage of terminal states during the evolution for each method, where the terminal state variables used are described in \autoref{sec:experiments}. We note that, for this experiment, $\varphi=0.0$, meaning NS-ES, Curiosity-ES, and TD3-ICM are motivated only by exploration. We do this to study exploration independently as an objective, but found similar results when using $\varphi=0.8$.

Curiosity-ES consistently explores more of the terminal states than other methods by the end of evolution, with the exception of the Stacker environment where MAP-Elites has a similar coverage distribution. As in the maze navigation task, this demonstrates the utility of Curiosity as it reaches high levels of exploration in terminal states without the definition of a behavior descriptor encouraging diversity in terminal states. MAP-Elites does well on all three environments, which we assume is related to the use of the same metric for behavior and coverage: reward-focused features of the terminal state. Suprisingly, CMAME has consistently lower exploration than MAP-Elites. We hypothesize that this is due to the emphasis in CMAME on exploitation; the CMA-ES emitter will spend many evaluations trying to improve the fitness of individuals. In the sparse setting, this results in a random walk when all fitness values are zero.

%We can see that at end Curiosity-ES is able to cover a greater space of the environment than the others algorithms on two of the benchmarks. This shows a great capacity of the algorithm to explore while the search for new states is not expressively defined in the intrinsic fitness. 
%Furthermore, we see that MAP-Elites outperform the coverage of Curiosity-ES in the STACKER benchmark. We think that this is due to the fact that we used, for the coverage, the exact same metric than for the behavior while in this environment a better description of the coverage would also take into account the position of the box. 
%Also the native mechanism of Curiosity is to focus in the direction of new dynamics. Therefore in STACKER it's likely that has soon has the dynamic of the gripper position has been appreciated, the algorithm will focus on harder dynamics such as throwing the box or playing with it ending in the stop of the maximisation of the gripper coverage space.
%Overall we see that MAP-Elites is extremely powerfull to explore a behavioral space when the behavioral space is not highly constrained. Indeed, here, wherever the gripper ends up, finding a new behavior from the previous one is really likely to happen. 

%Curiosity, however, is calculated on the entire trajectory and doesn't require the definition of a behavior descriptor, allowing it to find rewarding policies on all three tasks. On the Finger and Stacker tasks, Curiosity-ES also finds rewarding policies quickly and improves them to high levels of efficiency, solving the Finger task in less than 100 time steps.

% \begin{figure}[!h]
%     \centering
%     \textbf{Reward on Control Tasks}\\
%     \includegraphics[width=0.32\textwidth]{figure/FINGER_reward.pdf}
%     \includegraphics[width=0.32\textwidth]{figure/Ball_in_cup_reward.pdf}
%     \includegraphics[width=0.32\textwidth]{figure/Stacker_reward.pdf}
%     \caption{The sum reward of the best policy in each generation on the three control environments.}
%     \label{fig:control_reward}
% \end{figure}

Curiosity is able to lead an ES to reward in both the maze navigation and control tasks, finding rewarding policies on all six environments. Furthermore, Curiosity-ES is able to improve on the reward-finding policies, discovering more efficient policies than NS-ES on all environments. We posit that, compared to Novelty, Curiosity leads evolutionary search to areas where multiple transitions in a trajectory are different from previously observed transitions, driving the ES towards novel dynamics and areas of the environment. Finally, while the performance of other population based methods may depend on the choice of behavior descriptor, Curiosity-ES does not require any such definition and instead finds novel transitions based on Curiosity. We expand on the limitations of the behavior descriptor next.

\subsection{The bottleneck of the behavior descriptor}
\label{sec:bottleneck}

The exploration mechanism used by population based methods often require a behavior descriptor. While this behavior may be effective when appropriately defined, its definition requires expert knowledge and is not always evident. In order to explore the impact of the behavior descriptor, we expand the experimentation of the previous sections with a comparison to Novelty-based methods using as behavior the entire final state, which we term $BD_s$. We call these variations $\text{NS-ES\_BD}_{s}$ and $\text{NSLC\_BD}_{s}$. We note that, due to the expanded size of the final state, using the same discretization for MAP-Elites and CMAME would be overly computationally costly due to the increase in behavior dimension. We therefore focus on NS-ES and NSLC.

We also present a comparison to AURORA, which, like Curiosity-ES, does not require an explicit behavior descriptor. Instead, both methods use neural networks to analyze individual transitions over the full trajectory; for AURORA, states are used in order to define a behavior descriptor, and for Curiosity-ES, transition Curiosity. Unlike in \cite{cully_autonomous_2019}, we use a subsampling of the full trajectory to the use of 40\% of transitions. This is done to reduce the computational cost of the method, which is significantly more costly than Curiosity-ES due to the variable number of training epochs done per generation. Despite trying various hyperparameters such as subsampling rate and autoencoder learning rate, we were unable to find an instance of the AURORA method which found rewarding policies on any of the six environments.

\begin{figure}[!h]
    \centering
    \textbf{Coverage }\\
     \raisebox{0.8cm}{\includegraphics[width=0.10\textwidth]{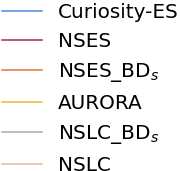}}
    \includegraphics[width=0.29\textwidth]{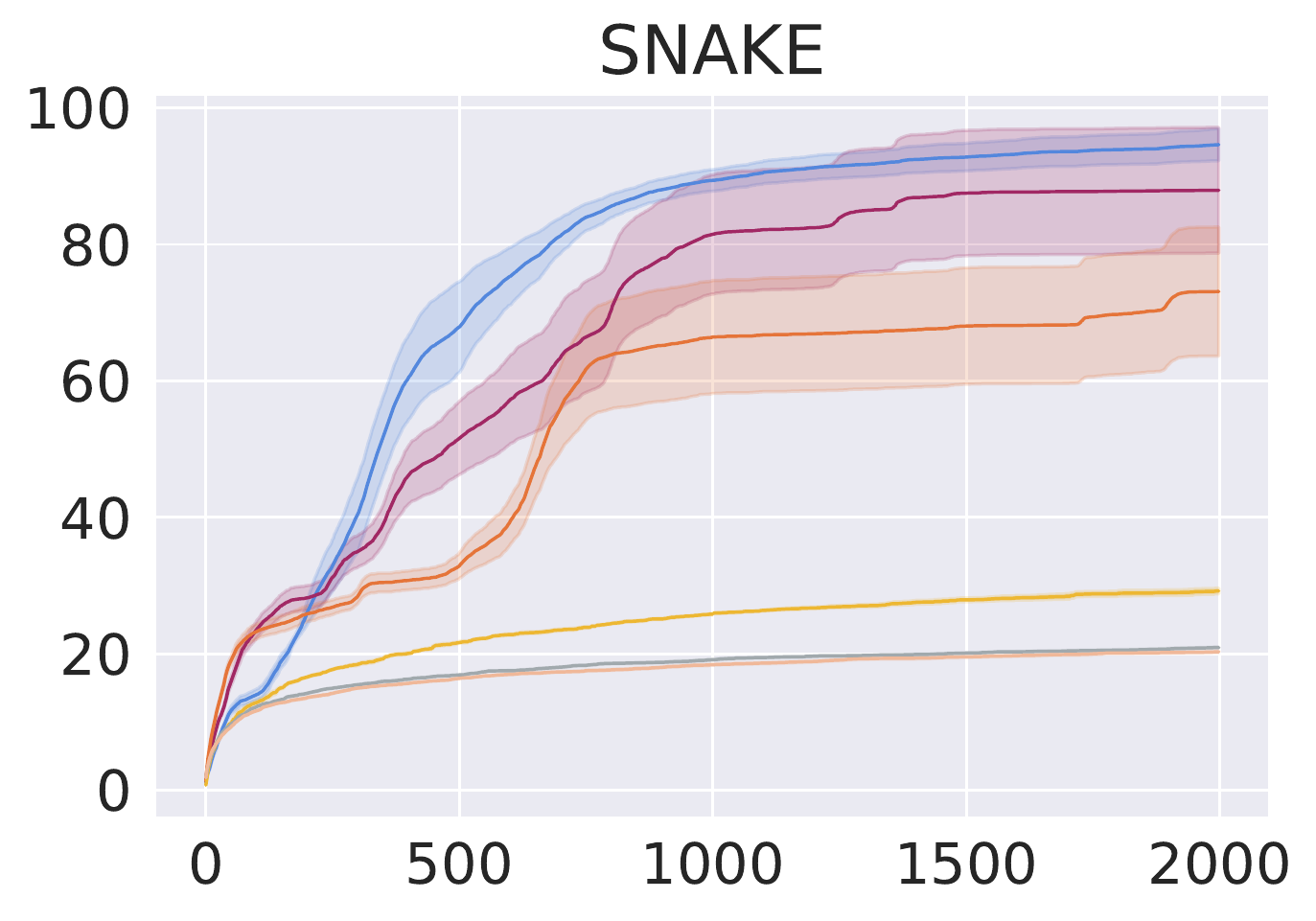}
    \includegraphics[width=0.29\textwidth]{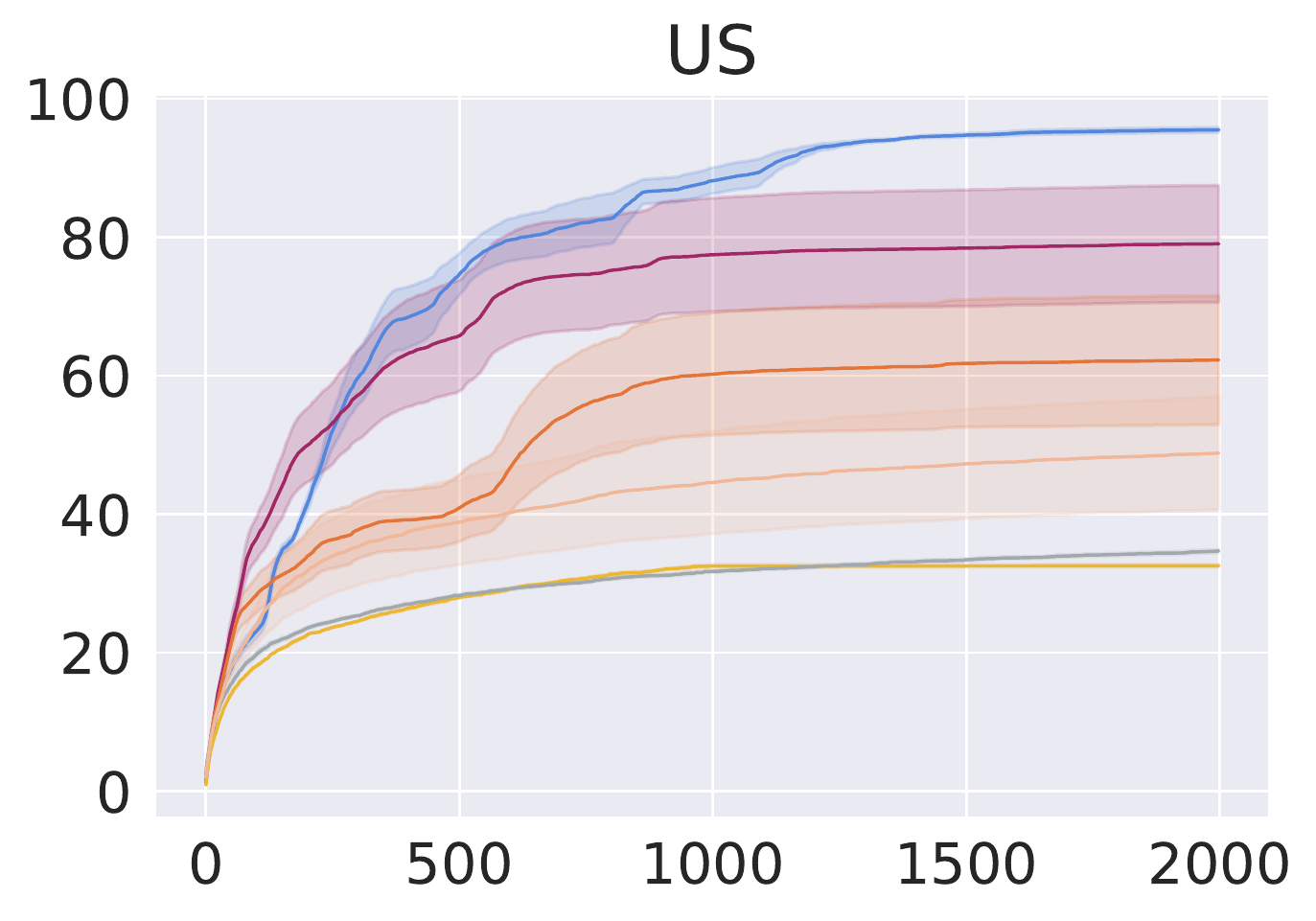}
    \includegraphics[width=0.29\textwidth]{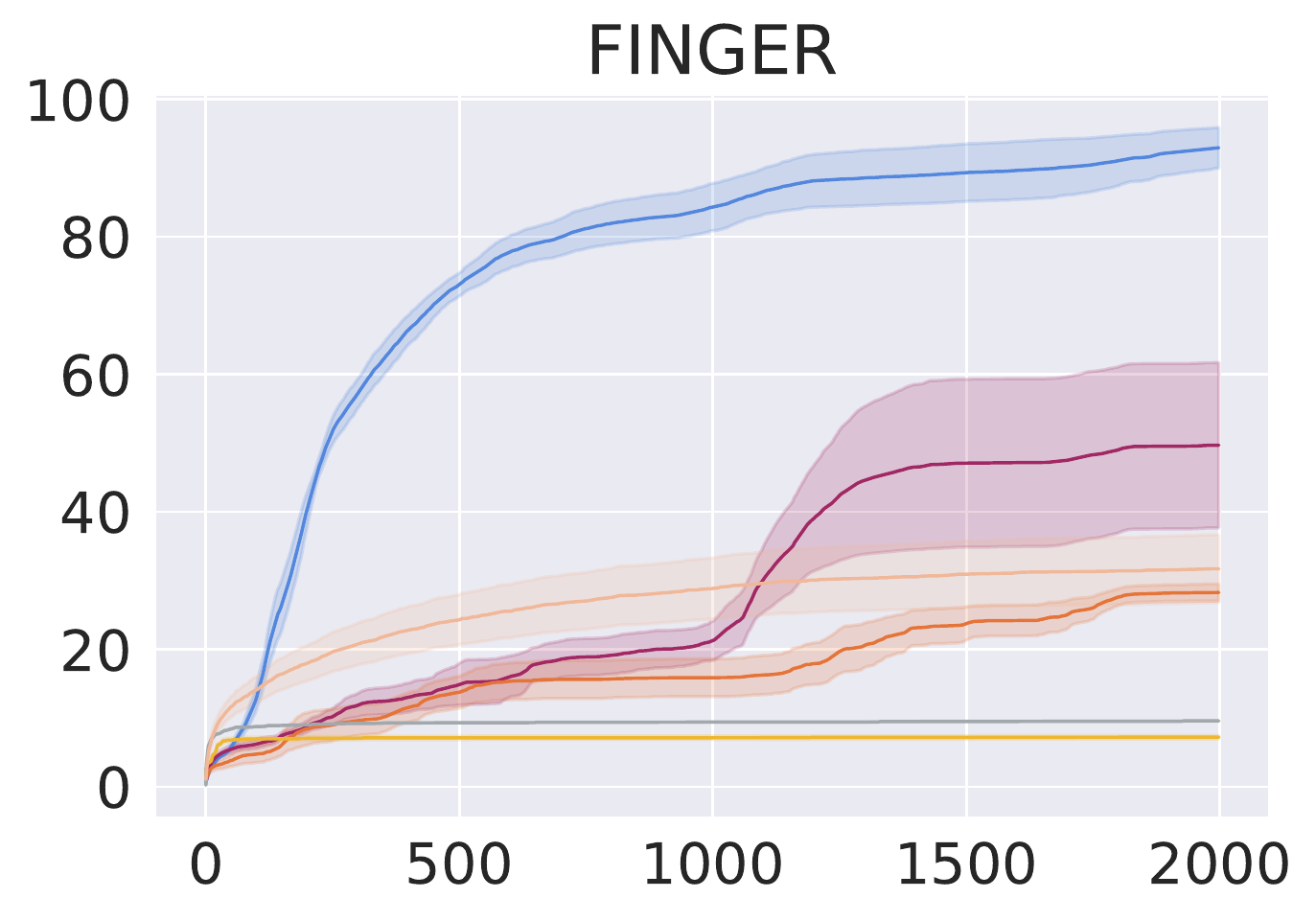}
    \caption{The percentage of discrete terminal states reached throughout policy search. For SNAKE and US, coverage measures final robot position, and for Finger, it measures the final finger joint and hinge positions.}
    \label{fig:control_coverage_sec}
\end{figure}

In \autoref{fig:control_coverage_sec}, we show the terminal state coverage of Curiosity-ES, NS-ES, AURORA, and NSLC on three demonstrative tasks: the SNAKE and US mazes and the Finger DMCS task. Results on the HARD maze and Ball in Cup and Stacker control tasks demonstrated similar results. We first note that NS-ES explores more than NSLC on all three tasks; as mentioned above, we believe that this is due to informed mutation of the ES, which is able to precisely update the center of the population distribution through gradient estimation. The coverage of NSLC is also inferior to that of MAP-Elites on all tasks. Neither NSLC nor AURORA found rewarding policies on any of the six tasks.

For both NS-ES and NSLC, the use of the full state as the behavior descriptor decreases coverage, where coverage is measured using the task-related features (e.g., position for the mazes). As such, Novelty based on the full state may be rewarded to individuals which do not increase coverage, making search for final position coverage less efficient. More concerning, however, is the difference in the US maze, where $\text{NS-ES\_BD}_{s}$ appears to converge to a lower total coverage than NS-ES. The inclusion of non-task information in the behavior descriptor can therefore prevent exploration under certain conditions. This effect is visible for NSLC as well, although less pronounced.

Curiosity-ES can also search in areas of the transition space which are not related to the task, such as exploring different actuator dynamics in the DMCS control suite. However, despite not providing task information in the form of a behavior descriptor, Curiosity-ES is able to maximize reward and exploration, measured as coverage, consistently across tasks. In the next section, we explore how this is achieved.

%As we can see on \autoref{fig:control_coverage_sec} showing the coverage on two mazes benchmark and one DMCS, there is a gap between the two behavior descriptor. This is really highlighted with NS-ES schema where the 'optimal' behavior reaches approximately 15\% more of the coverage in each of the environments. Furthermore we can see that NSLC didn't cover appropriately the state space in the maze suit. Bringing us to the same conclusion made earlier. For highly constrained environments an informational mutation provided by an ES, augmented with an intrinsic motivation is really beneficial to the exploration. We believe that our implementation of AURORA didn't quite explore the state space due to the up-sampling of the trajectory. However, this result shows the limitations of such a method if it requires to store the full trajectory of all individuals.
%Overall, Curiosity-ES allows to explore more effectively the state space of the environment the the other methods while not having to hand-designed a behavior descriptor.

% The descriptor affects hugely the performance of the algorithms

\section{Diversity among the rewarding policies}
\label{sec:diversity_rewarding_policy}

%The field of Genetic programming has often studied the idea of multi-modal solutions. In our benchmarks this maybe translate as the seek for different rewarding policies.

In the previous section, we demonstrated that Curiosity-ES is able to find efficient rewarding policies while exploring large parts of the transition space. To better understand how this is done, we explore the full set of rewarding policies generated over an evolution. We use Principal Component Analysis (PCA) to create a two-dimensional representation of the rewarding policies generated by each algorithm. This allows for a visual comparison and evaluation of the performance of each method. To do so, we define the feature space for each rewarding policy as the last 300 states sampled by the policy in the environment. We based this analysis on the maze US, as CMAME, MAP-Elites, and Curiosity-ES all had high reward on this task, and CMAME and Curiosity-ES had nearly full coverage.

\begin{figure}[!h]
    \centering
    \textbf{PCA over the 300 last states}\\
    \includegraphics[width=\textwidth]{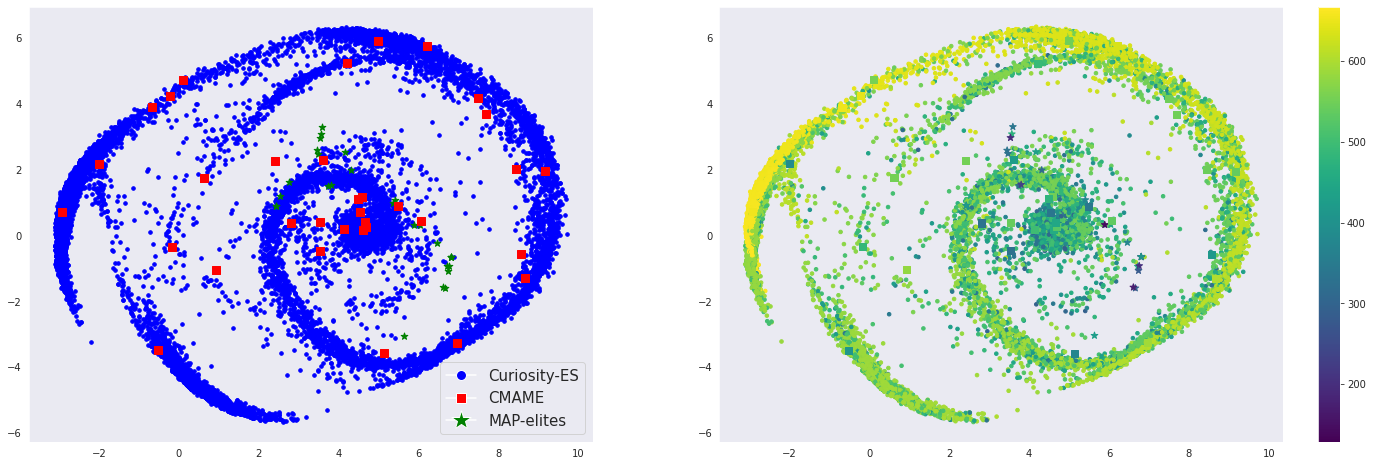}
    \caption{Projection in a two dimensional space of the 300 last states for each policy that found the reward in the US environment. On the left is the representation of the projection, colored by algorithm. On the right is the same projection, colored by the extrinsic fitness of the policy.}
    \label{fig:pca}
\end{figure}

The results are shown in \autoref{fig:pca}. In this figure, each point represents a projection of a policy that found the reward in the maze US. We first note that MAP-Elites was only able to sample a small number of distinct rewarding policies while not exploring sufficiently around these policies in order to maximize the fitness. This leads to a highly partial space coverage of the subspace of reward policies. CMAME was able to sample a variety of rewarding policies with very different behavior over the last 300 states. However, similar to MAP-Elites, there is little exploration in the neighborhood of each policy, making rewarding policies very different from another.

Curiosity-ES is able to explore continuously throughout the space of rewarding behaviors, covering the policy areas found by CMAME and MAP-Elites but finding many other similar rewarding policies. This allows for optimization of the extrinsic reward around each of the behavior centers discovered. While CMAME is able to find highly rewarding policies on this problem, it finds few and only using certain behaviors (the points near the origin of the PCA). Curiosity-ES is able to find a multitude of highly rewarding policies in different parts of the behavior space.

%On the opposite CMAME which makes the use of CMA-ES in order to maximize the extrinsic fintess was able to evolve its path in the policy space in order to sample the most rewarding policies. This induce a more optimal coverage of the subspace of rewarding policies. 
%However we can see that Curiosity-ES demonstrated a much more uniform coverage of the subspace of rewarding policies, effectively sampling policies beyond the specific distribution while still including the most rewarding policies. We can see how well the algorithm is able to extract the topology of the subspace of rewarding policies. 
%We think that this analysis could lead to new approach where the seek for multi-modal solutions is not explicitly formalised during the evolution but rather found with these type of analysis afterwards.

\section{Discussion}
\label{sec:discussion}

% summarize contribution

In the article we show that Curiosity is an effective way to promote policy exploration in population-based methods. We demonstrate this on maze navigation and robotic control tasks and show high levels of coverage and multiple reward reaching policies on all tasks using an evolutionary strategy with Curiosity as intrinsic fitness. We posit that Curiosity naturally leads to a diversity of policies due to the reward associated with covering unexplored or under-explored transitions.

% comparison to NS
We base our comparison on existing population-based methods in Quality Diversity due to the similarity of Curiosity to other diversity metrics such as Novelty used in QD algorithms. We chose simple environments of types that NS has already been demonstrated on, namely maze navigation, and observed that Curiosity was able to compete with or outperform all other QD methods on all tasks. However, Curiosity was originally explored with image-based environments \cite{pathak2017curiosity} and the feature encoding module $\phi$ was conceived to allow for scaling to problems of large input dimension. While a comparison with other QD methods on image-based tasks would be complicated by the necessity to define a behavior descriptor, we aim to study Curiosity-ES on image-based tasks in future work.

The native mechanism of Curiosity-ES is to focus search towards new dynamics; in the maze navigation task, this means maze sections which are different in form, and in the DMCS tasks, this means transitions which display a different dynamic of interaction between the various objects. In the Stacker setting, where a robotic arm stacks boxes at a target location, novel dynamics may be explored by behaviors such as throwing boxes or rotating the gripper joints. These behaviors are not captured by the behavior descriptor and coverage measure, which was the final position of the gripper in our experiments. Curiosity-ES does not explore all terminal positions of the gripper as the dynamics of the movement of the gripper may be learned early by the ICM. MAP-Elites, however, is able to explore new final gripper positions consistently as the behavior descriptor explicitly encourages that exploration. While the definition of a behavior descriptor is often a challenging aspect of QD algorithm application, for applications where the exploration of a specific and quantifiable measurement is desirable, a combination of Curiosity-ES and MAP-Elites could be pertinent and represents a possible future direction.

The ICM is motivated by the idea that neural networks can accurately model the transitions already covered but generalize poorly to unseen data, i.e. that the prediction on new transitions will have high error. Curiosity is therefore suited for exploring environments where the transitions change sufficiently to incur error; for example, in the SNAKE maze, the repetitive sections of the maze did not incur high error due to their similarity to previous transitions, even when newly discovered. The learning rate of the ICM is therefore an important hyperparameter, as quick convergence may discourage exploration of similar areas. The use of other exploration bonuses, as in RND \cite{burda2018exploration}, DIAYN \cite{eysenbach2018diversity}, or Never Give Up \cite{badianever}, could also be an interesting direction for population-based policy search.

% combinations with RL
As exploration is more difficult in gradient-based RL than in population-based methods, there are many opportunities to combine benefits from these two domains. We show that Curiosity, originally demonstrated in RL, can be very effective in an ES. Similarly, Novelty has been used to encourage exploration in RL methods \cite{shi2020efficient, liu2021pns}. Other ideas from RL such as the adversarial training schemes in \cite{flet2020adversarially} and pink noise exploration in \cite{eberhard2022pink} could be adapted to ES. Curiosity is evolved in \cite{alet2020meta}; a similar meta-learning or meta-evolutionary approach could be taken in Curiosity-ES. \cite{sigaud2022combining} provides a review of methods which combine RL and population-based methods; as demonstrated with Curiosity, we believe there is a great potential in this intersection.

% final paragraph
We found in this work that exploration bonuses which reward transitions that are different leads to the evolution of a diverse set of policies. This motivates this idea of using exploration bonuses which are calculated throughout the lifetime of an individual, rather than basing reward on an aggregate measure of behavior. We observed that the use of an intrinsic motivation over the entire trajectory led to more efficient policies and greater overall exploration.

\printbibliography

\end{document}